\PassOptionsToPackage{numbers,compress}{natbib} 
\documentclass{article}

\usepackage[final]{neurips_2025}
\usepackage{graphicx}

\usepackage[utf8]{inputenc} 
\usepackage[T1]{fontenc}    
\usepackage{hyperref}       
\usepackage{url}            
\usepackage{booktabs}       
\usepackage{amsmath}        
\usepackage{amsfonts}       
\usepackage{nicefrac}       
\usepackage{microtype}      
\usepackage{xcolor}         

\title{Binary Sparse Coding for Interpretability}

\author{%
  Lucia Quirke \\
  EleutherAI\\
  \texttt{lucia@eleuther.ai} \\
  \And
  Stepan Shabalin \\
  Georgia Institute of Technology\\
  \AND
  Nora Belrose \\
  EleutherAI \\
}

\begin{document}

\maketitle

\begin{abstract}

Sparse autoencoders (SAEs) are used to decompose neural network activations into sparsely activating features, but many SAE features are only interpretable at high activation strengths. To address this issue we propose to use binary sparse autoencoders (BAEs) and binary transcoders (BTCs), which constrain all activations to be zero or one. We find that binarisation significantly improves the interpretability and monosemanticity of the discovered features, while increasing reconstruction error. By eliminating the distinction between high and low activation strengths, we prevent uninterpretable information from being smuggled in through the continuous variation in feature activations. However, we also find that binarisation increases the number of uninterpretable ultra-high frequency features, and when interpretability scores are frequency-adjusted, the scores for continuous sparse coders are slightly better than those of binary ones. This suggests that polysemanticity may be an ineliminable property of neural activations.

\end{abstract}
    
\section{Introduction}

Recently, large language models have reached human‐level reasoning across numerous tasks \citep{guo2025deepseek}. The field of interpretability seeks to bolster these models' safety and dependability by elucidating their internal structures and representations. Although early efforts focused on generating natural‐language descriptions for individual neurons \citep{Olah2020,gurnee2023finding,gurnee2024universal}, it is now commonly understood that the majority of neurons are ``polysemantic,'' responding to a variety of semantically unrelated inputs \citep{arora2018linear,elhage2022toy}.

Sparse autoencoders (SAEs) have recently shown promise in mitigating polysemanticity by decomposing activations into more coherent, interpretable components \cite{bricken2023monosemanticity,templeton2024scaling,gao2024scaling}. An SAE is a single‐hidden‐layer network trained to reconstruct its input activations while enforcing sparsity through penalties \cite{bricken2023monosemanticity,rajamanoharan2024jumping}, explicit constraints \cite{gao2024scaling,bussmann2024batchtopksparseautoencoders}, or an information‐bottleneck objective \cite{ayonrinde2024interpretability}. The architecture comprises an encoder that maps activations into a sparse, overcomplete latent representation, and a decoder that reconstructs the original activations from this latent code.

In general, SAE features are qualitatively more interpretable than neurons. But many SAE features are uninterpretable at low activation strengths, while interpretability evaluations often focus on high activation strengths. There is a general concern that SAEs might ``hide'' uninterpretable information in the exact activation strength of each active feature. In this work, we address these issues by \emph{binarising} SAE activations, constraining them to take values in the set $\{0, 1\}$. We develop a sigmoid-based straight through estimator (STE) for the gradient \citep{bengio2013estimating}, allowing us to differentiate through the discontinuous binarisation operation.

\section{Related work}

\citet{tamkin2023codebook} finetune neural network layers to add sparse binary bottlenecks with a similar structure to sparse autoencoders, and finds that the resulting features are highly interpretable and can be used to steer the model. Unlike sparse autoencoders, codebooks are components of the base model, use features' cosine similarities with the input rather than their dot products to select the top $k$, tie encoder and decoder weights, and do not use bias terms.

\citet{ayonrinde2024interpretability} quantise SAE activations into discrete bins after training for the purpose of encoding the data in the most efficient way possible. Instead of binarising they perform 8-bit quantisation.

\citet{gallifant2025classification} use an SAE feature binarisation step in their inference-time sequence classification pipeline; they set non-top-n feature activations to zero at each token, sum the feature vectors, then binarise each component of the result with a fixed threshold of 1.

\section{Preliminaries} 

Sparse autoencoders consist of two parts: an encoder that transforms activation vectors into a sparse, higher-dimensional latent space, and a decoder that projects the latents back into the original space. Different activation functions can be used in the encoder, but in this work we will focus on the TopK function introduced by \citet{gao2024scaling} due to its simplicity and efficacy. It zeros out all activations which are not in the top $k$ highest activations on a given token, thereby directly enforcing a given level of sparsity. The functional form of a TopK SAE is thus
\begin{equation}
    \boldsymbol{\hat{x}} = \mathbf{W}_{\mathrm{dec}} \mathrm{TopK}(\mathbf{W}_{\mathrm{enc}} \boldsymbol{x} + \mathbf{b}_{\mathrm{enc}}) + \mathbf{b}_{\mathrm{dec}}
\end{equation}
All parameters are trained jointly to minimise the reconstruction error $||\boldsymbol{x} -\boldsymbol{\hat{x}}||^2_2$. It is standard to initialise $\mathbf{W}_{\mathrm{dec}}$ and $\mathbf{W}_{\mathrm{enc}}$ to be transposes of each other, except that the rows of $\mathbf{W}_{\mathrm{dec}}$ are constrained to be unit-norm \citep{bricken2023monosemanticity}. We also initialise $\mathbf{b}_{\mathrm{dec}}$ to the empirical mean or geometric median of the MLP outputs. This allows the fraction of variance explained to start out significantly higher than zero.

\begin{figure*}[t]
    \centering
    \includegraphics[trim=0 1cm 0 1cm, clip, width=1.\textwidth]{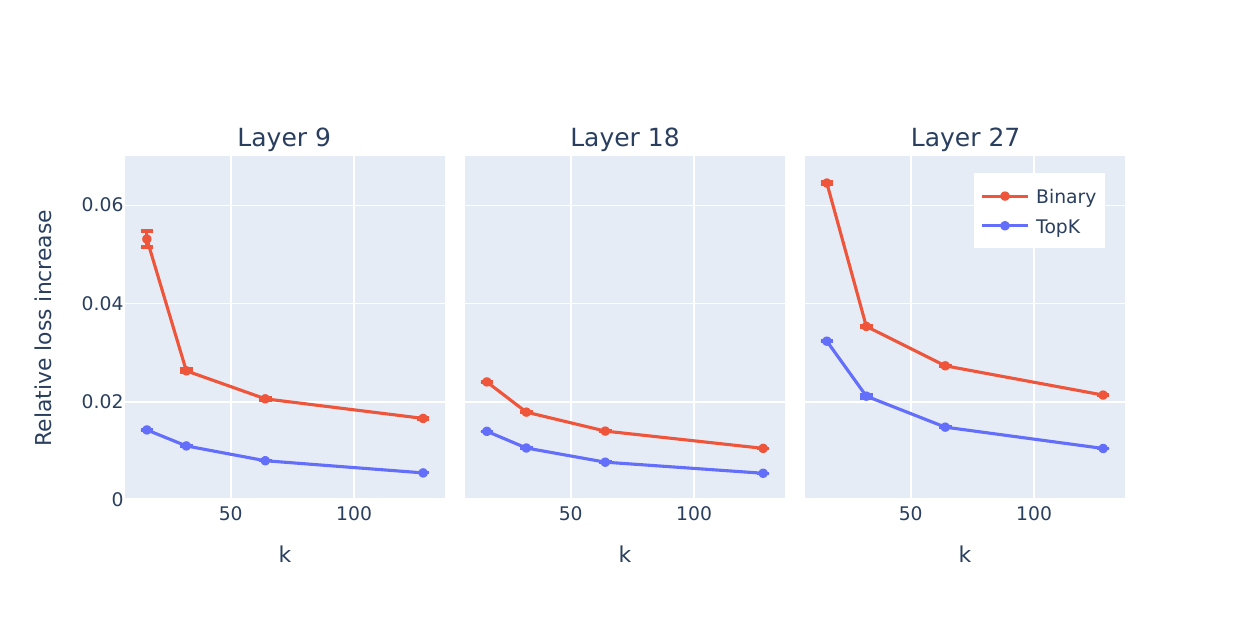}
    \caption{Binary vs. continuous next token prediction cross-entropy loss increase over various values of $k$. Each model are trained on 20 billion tokens.}
    \label{fig:loss_increase}
\end{figure*}
\begin{figure*}[t]
    \centering
    \includegraphics[trim=0 1cm 0 1cm, clip, width=1.\textwidth]{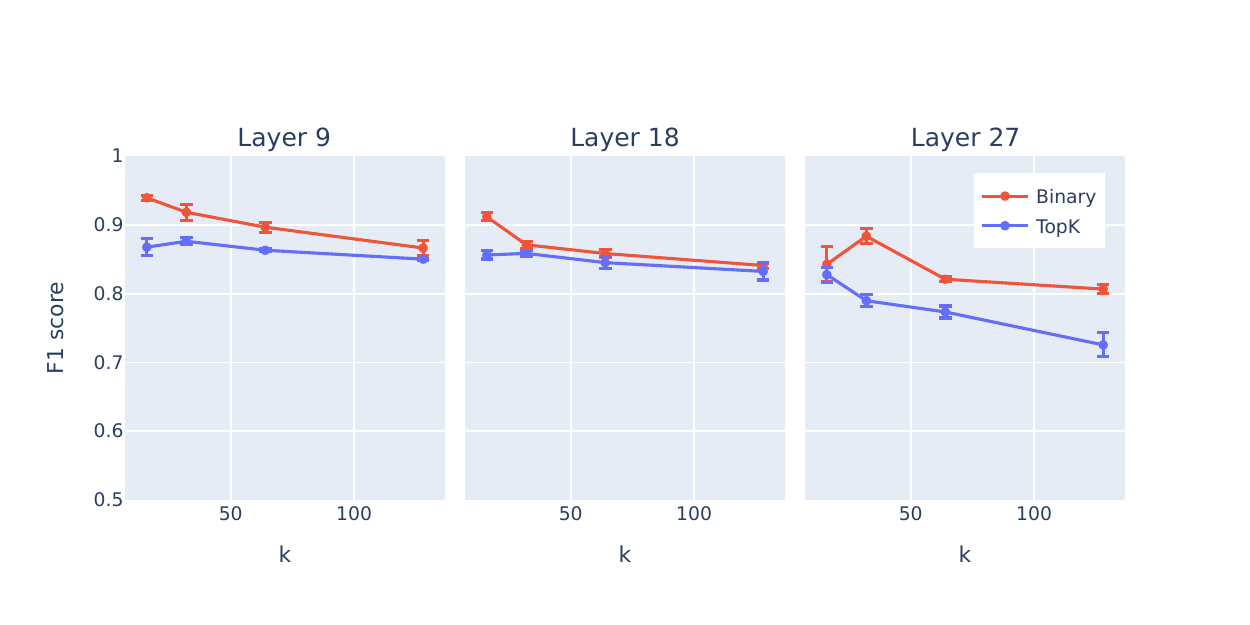}
    \caption{Unweighted fuzzing interpretability scores for binary vs. continuous skip-transcoders trained on SmolLM2-135M, with various values of $k$. By this metric, binary coders match or outperform continuous ones across all layers and values of $k$.}
    \label{fig:btc_f1}
\end{figure*}
\begin{figure*}[t]
    \centering
    \includegraphics[trim=0 1cm 0 1cm, clip, width=1.\textwidth]{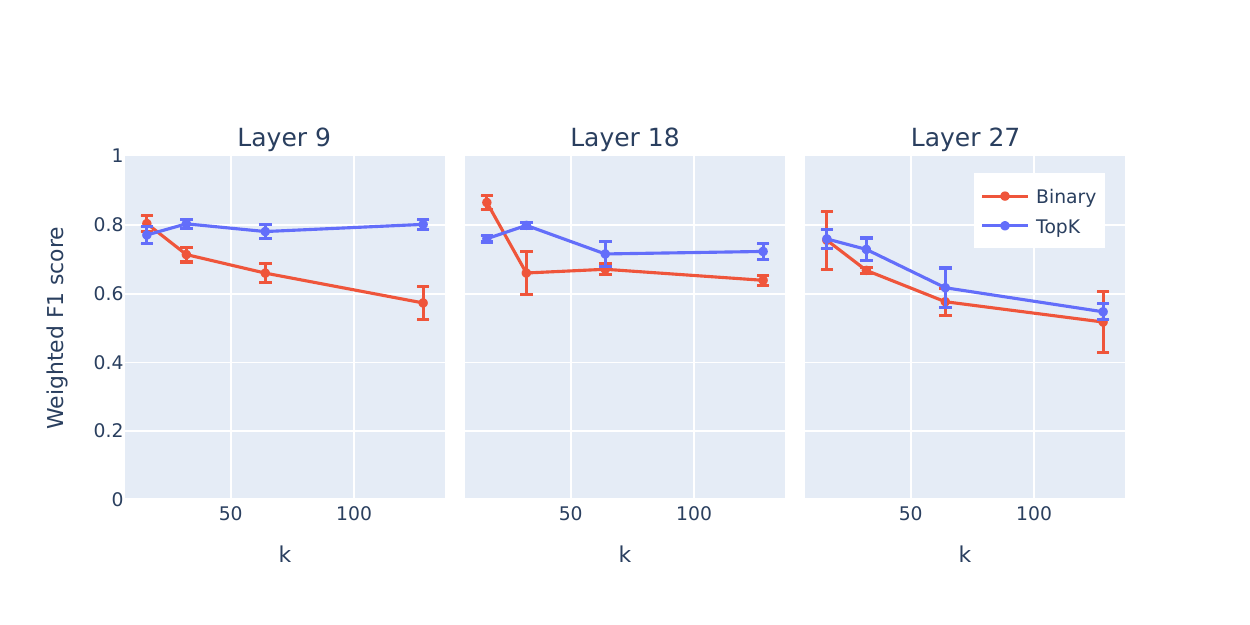}
    \caption{Frequency-weighted fuzzing interpretability scores for binary vs. continuous skip-transcoders trained on SmolLM2-135M, with various values of $k$. By this metric, binary coders are often less interpretable than continuous ones.}
    \label{fig:weighted_interp} 
\end{figure*}

Later, \citet{dunefsky2024transcoders} proposed \emph{transcoders} as an alternative method for extracting interpretable features from transformers. The architecture of a transcoder is identical to that of an SAE, but its target output is different---given the input of a feedforward component such as an MLP or attention block, a transcoder is trained to predict the component's output. For transcoders we do not constrain $\mathbf{W}_{\mathrm{dec}}$ to have unit-norm rows, and because we are no longer trying to approximate an identity function $\mathbf{W}_{\mathrm{dec}}$ is zero-initialised rather than being initialised with the transpose of $\mathbf{W}_{\mathrm{enc}}$. We also initialize $\mathbf{b}_{\mathrm{enc}}$ to the inverse of the empirical mean of the encoder $-\mathbf{W}_{\mathrm{enc}} \mathbb{E}[\boldsymbol{x}]$ to ensure preactivations are centered at initialization.. The transcoder starts out training as a constant function with fraction of variance explained equal to zero.

There exist transcoder variants that introduce additional parameters for the sake of increasing reconstruction accuracy. \cite{ameisen2025circuit} adds decoder weights that connect residual streams at different layers, and \citet{paulo2025transcoders} shows that the simple modification of allowing a linear skip connection between the inputs and outputs of a transcoder improves reconstruction performance at no cost to interpretability scores. The \textbf{skip-transcoder} takes the functional form
\begin{equation}
    \boldsymbol{\hat{y}} = \mathbf{W}_{\mathrm{dec}} \mathrm{TopK}(\mathbf{W}_{\mathrm{enc}} \boldsymbol{x} + \mathbf{b}_{\mathrm{enc}}) + \mathbf{W}_{\mathrm{skip}} \boldsymbol{x} + \mathbf{b}_{\mathrm{dec}}
\end{equation}
where $\boldsymbol{x}$ is the input of the MLP and $\hat{y}$ is the reconstructed output of the MLP. The $\mathbf{W}_{\mathrm{skip}}$ parameter is zero-initialised, so that skip-transcoders start out as constant functions just like normal transcoders.

Importantly, the work found that transcoders and skip-transcoders yield features with higher interpretability scores on average than SAEs. While transcoders are sometimes worse than SAEs in terms of reconstruction loss, the skip connection fixes this issue. For these reasons, we emphasise skip-transcoders in this work. Where our discussion refers to multiple model types, we will use the general term \textit{sparse coder}.

\section{Methods}

\subsection{Binary sparse coders}

Binary sparse coders are a simple modification of standard sparse coders. We simply add a binarisation step after the TopK function is applied, forcing the activations to binary as well as sparse. The functional form for a binary SAE is thus
\begin{equation}\label{eq:bae}
    \boldsymbol{\hat{y}} = \mathbf{W}_{\mathrm{dec}} \mathrm{Binarise}[\mathrm{TopK}(\mathbf{W}_{\mathrm{enc}} \boldsymbol{x} + \mathbf{b}_{\mathrm{enc}})] + \mathbf{b}_{\mathrm{dec}}
\end{equation}
where $\mathrm{Binarise} : \mathbb{R}^N \rightarrow \mathbb{R}^N$ replaces each element of a vector with $1$ if it is positive, otherwise replacing it with $0$. The sigmoid function can be viewed as a continuous and differentiable relaxation of $\mathrm{Binarise}$, and hence we use a sigmoid-based straight-through estimator \citep{bengio2013estimating} to backpropragate through it. We found that adding a temperature parameter of $2$ to the sigmoid STE improved training stability.

\subsection{Gumbel-Softmax and GroupMax}

Reconstruction accuracy can be further improved by training with the Gumbel-Softmax gradient estimator \citep{jang2017categorical}, which enables differentiable approximate categorical sampling of pre-activations through a relaxation of the Gumbel-Max trick. We found that this method can create training instability when backpropagating through the top-$k$ operator, so for the Gumbel variant we replaced the TopK with a novel activation function we call \emph{GroupMax}, that groups latents into $k$ groups and selects the largest element in each group. These \emph{GroupMax} sparse coders reliably trained with Gumbel-Softmax without instability. The functional form for a binary GroupMax SAE is
\begin{equation}\label{eq:bae_gm}
    \boldsymbol{\hat{y}} = \mathbf{W}_{\mathrm{dec}} \mathrm{Binarise}[\mathrm{GroupMax}(\mathbf{W}_{\mathrm{enc}} \boldsymbol{x} + \mathbf{b}_{\mathrm{enc}})] + \mathbf{b}_{\mathrm{dec}}
\end{equation}
where $\mathrm{GroupMax} : \mathbb{R}^N \rightarrow \mathbb{R}^N$ divides a vector into $k$ contiguous and equally sized groups, then zeros out any activation that is not the largest of a group.


\subsection{Training}

Unless stated otherwise, all experiments train sparse coders on a total of 10 billion tokens with a batch size of $2^{20}$ tokens. The training data is drawn from the base model's original corpus. For SmolLM2-1.7B, which was trained with curriculum learning, we specifically use the final-stage dataset mixture from its pre-training curriculum.

We trained skip-transcoders, binary skip-transcoders, and the Gumbel GroupMax skip-transcoder variant at $k=32$ for both SmolLM2-135M and SmolLM2-1.7B. 

We also trained skip-transcoders, binary skip-transcoders, SAEs, and BAEs  over a variety of $k$ values for the SmolLM2-135M base model. We trained three seeds for each sparse coder and report mean values for our evaluation metrics; error bars denote the $\pm\,$1 standard error of the mean.

We additionally trained three seeds of the Gumbel GroupMax variant at $k=32$, and finetuned it along with binary and continuous skip-transcoders of equal $k$ using a KL divergence loss for use in sparse probing.

We find that fine-tuning transcoders with a KL divergence loss as proposed by \citet{karvonen2025finetuneSAE} can improve performance when the transcoder is patched into the base model.

We use the Adam optimiser \citep{kingma2017adam} to train most sparse coders on SmolLM2-135M models, and the schedule-free Signum optimiser \citep{bernstein2018signsgd, defazio2024road} to train the SmolLM2-1.7B coders and to train binary SAEs, which exhibit near-complete index collapse when trained with Adam. For Adam we used a linear learning rate schedule with a peak value of $3\times 10^{-3}$ for transcoders, $5\times 10^{-3}$ for SAEs, 1000 warmup steps, $\beta_1=0.9$, $\beta_2=0.999$, and no weight decay. For schedule-free Signum we used a constant learning rate of $3\times 10^{-3}$ and a momentum of $0.95$. Note that in schedule-free training the momentum term is equivalent to the weight EMA employed by \citet{gao2024scaling}.

\begin{figure*}[t]
    \centering
    \includegraphics[trim=0 0 0 0, clip, width=0.95\textwidth]{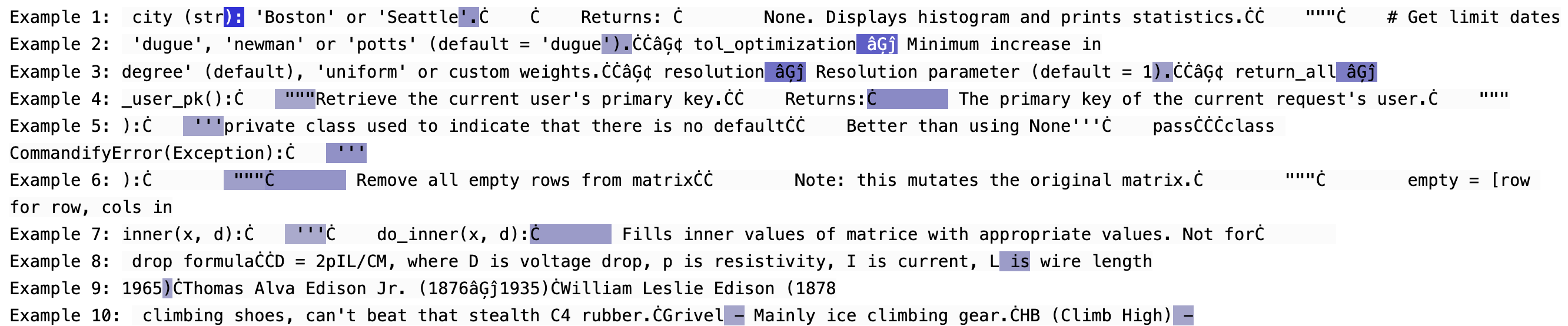}
    \caption{Activating examples from different deciles of a TopK skip-transcoder feature. The top activations appear in a coding context, while the lower activations appear in diverse contexts.}
    \label{fig:topk_low_acts}
\end{figure*}
\begin{figure*}[t]
    \centering
    \includegraphics[trim=0 0 0 0, clip, width=0.95\textwidth]{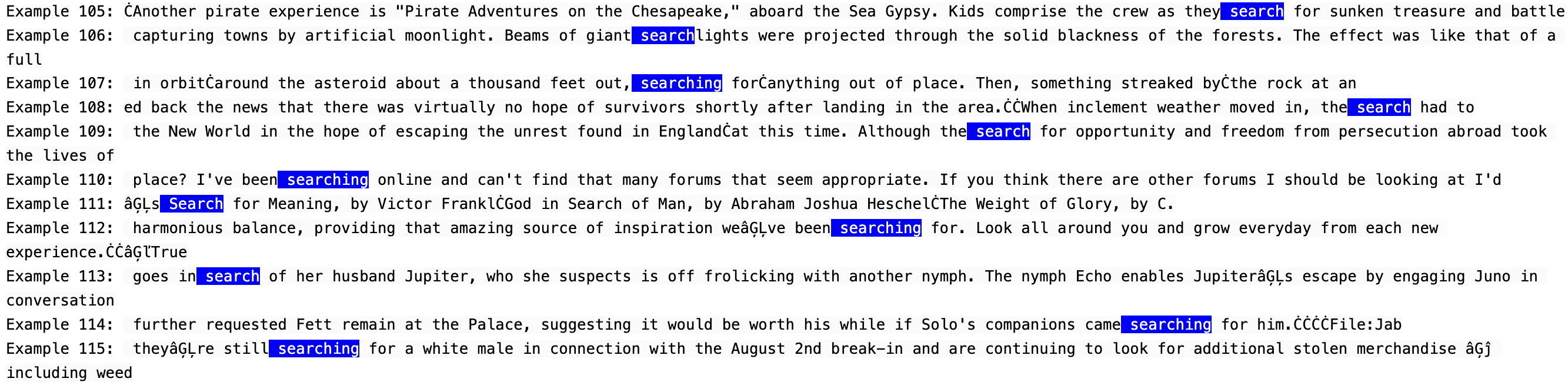}
    \caption{Activating examples for a binary skip-transcoder feature that achieved perfect auto-interpretability scores. The generated explanation is ``The verb `search' or its variants, often used in contexts of looking for something, someone, or information, and frequently associated with a sense of investigation, inquiry, or pursuit.'' Note that a counterexample to this explanation, a non-activating instance of ``Search'', appears mid-way through Example 111.}
    \label{fig:contrastive_examples}
\end{figure*}
\begin{figure*}[t]
    \centering
    \includegraphics[trim=0 0 0 0, clip, width=0.95\textwidth]{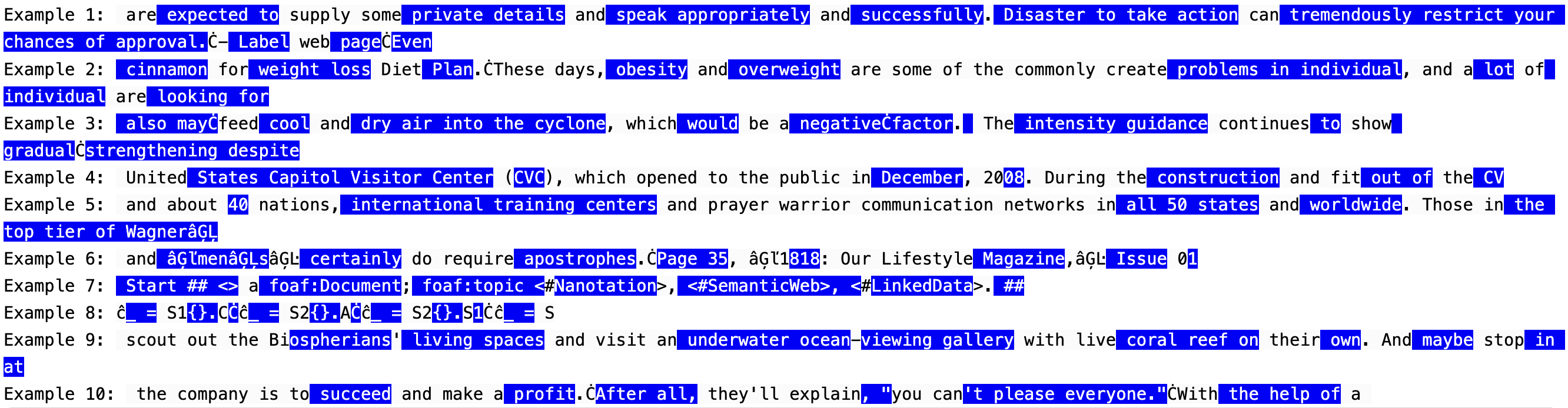}
    \caption{Activating examples from an ultra high frequency binary skip transcoder feature. The activations occurs on more than half of tokens and are not interpretable.}
    \label{fig:ultra_high}
\end{figure*}

\subsection{Evaluation}

Our primary performance metric is the increase in next-token cross-entropy loss induced by patching the sparse coder into its base model. We also provide the trajectory of the fraction of variance unexplained (FVU) training objective (Appendix~\ref{app:fvu}.)

To evaluate the interpretability of the sparse coders, we employ the auto-interpretability techniques proposed in \cite{paulo2024delphi} to generate and evaluate feature explanations using LLMs. First, we collect sparse coder feature activations over 10 million tokens from the training dataset, filtering out activations associated with beginning of sentence (BOS) tokens. Next, we sample 100 features from each of four layers of the sparse coder to explain, for a total of 400 features. We generate an explanation for each feature by prompting an LLM with 40 decile-sampled labeled activating sequences of 256 tokens each, then attempt to classify balanced datasets containing 50 sequences with decile-sampled feature activations and 50 randomly sampled non-activating sequences by prompting an LLM to classify each sequence as activating or not activating in accordance with the generated feature explanation. We use Llama 3.1 for both explanation generation and sequence classification.

The results of the LLM binary classification tasks can be used to produce the mean classification accuracy, but this metric can be misleading because sparse coder features tend to exhibit extreme class imbalance where they are inactive far more than they are active. To account for this imbalance we report the mean $F_1$ score instead. The $F_1$ score for a feature is the harmonic mean of the precision and recall, which simplifies to

\begin{equation}\label{eq:f1}
F_1 = \frac{2 \cdot \text{true positives}}{2 \cdot \text{true positives} + \text{false positives} + \text{false negatives}} 
\end{equation}

As proposed in \cite{paulo2024delphi}, we calculate scores for both detection and fuzzing classifiers, which perform slightly different classification tasks. For the detection task the classification task is to determine whether or not a sequence contains a feature activation, and for fuzzing the classification task is to determine whether a highlighted token in a sequence is associated with a feature activation.

We also provide the frequency-adjusted F1 scores for each classifier, since features vary in their firing frequencies across many orders of magnitude.

In addition to the aforementioned metrics, we integrate the sparse probing evaluation from \citep{karvonen2024saebench}. This evaluation tests representation of concepts in the SAE by training sparse classifiers of samples containing those concepts (Table~\ref{tab:sparse_probing}.)

\section{Results}

\begin{figure*}[t]
    \centering
    \includegraphics[trim=0 1cm 0 2cm, clip, width=0.49\textwidth]{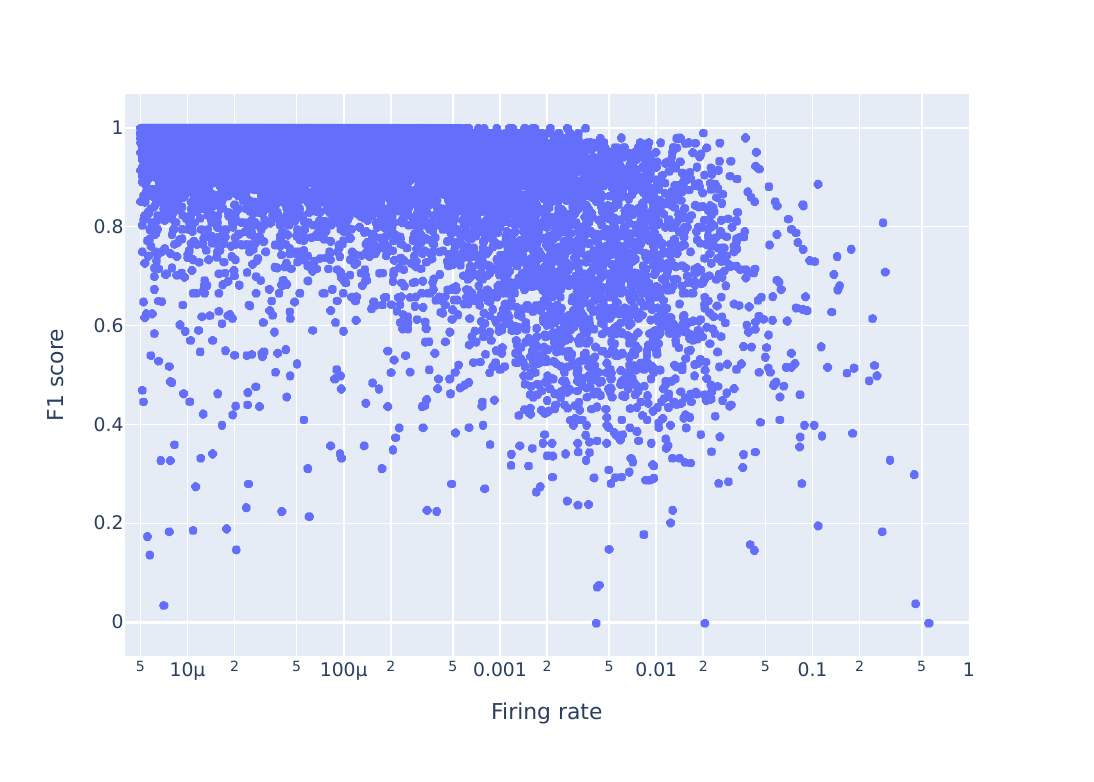}
    \includegraphics[trim=0 1cm 0 2cm, clip, width=0.49\textwidth]{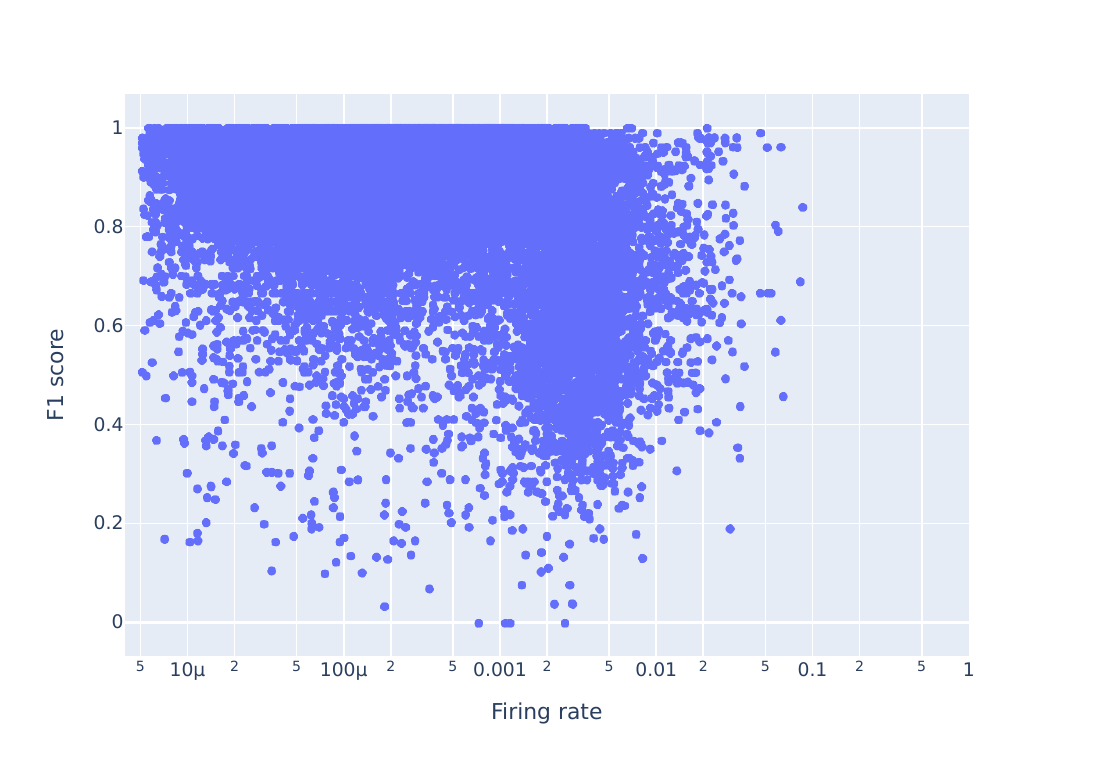}
    \caption{Feature firing rates for binary (left) and continuous (right) skip-transcoders on layer 9 of SmolLM2-135M, calculated over 10 million tokens. Firing rates are plotted on a log scale. The binary transcoder has  ultra-high frequency features than the continuous transcoder, with the most frequent firing on over half of tokens. No continuous feature activates on more than 10\% of the dataset.}
    \label{fig:firing_rates_comparison}
\end{figure*}

We find that binarisation improves unweighted average feature interpretability (Figure~\ref{fig:btc_f1}), but increases the cross-entropy loss when patched into the base model (Figure~\ref{fig:loss_increase}). Unfortunately, we find that binarisation does not increase interpretability when features are weighted by their firing frequencies. We find that binary transcoders indeed have more ultra-high frequency features than continuous ones do, and that these tend to have fairly low interpretability scores (Figure~\ref{fig:firing_rates_comparison}). We examine a concrete example of this phenomenon in Figure~\ref{fig:ultra_high}. We hypothesize that the sparse coder effectively learns to push the information that would usually be stored in low activation deciles into high-frequency, uninterpretable features. If this is true, it would suggest that polysemanticity is to a certain extent ineliminable: try to remove it, and it gets pushed somewhere else.

Binarised models also underperformed continuous ones in the sparse probing evaluation, however after finetuning both models the opposite is true; both binary skip-transcoders and their Gumbel variant outperformed finetuned continuous skip-transcoders (Table~\ref{tab:sparse_probing}.)

\begin{figure*}
    \centering
    \includegraphics[trim=0 1cm 0 1cm, clip, width=0.7\textwidth]{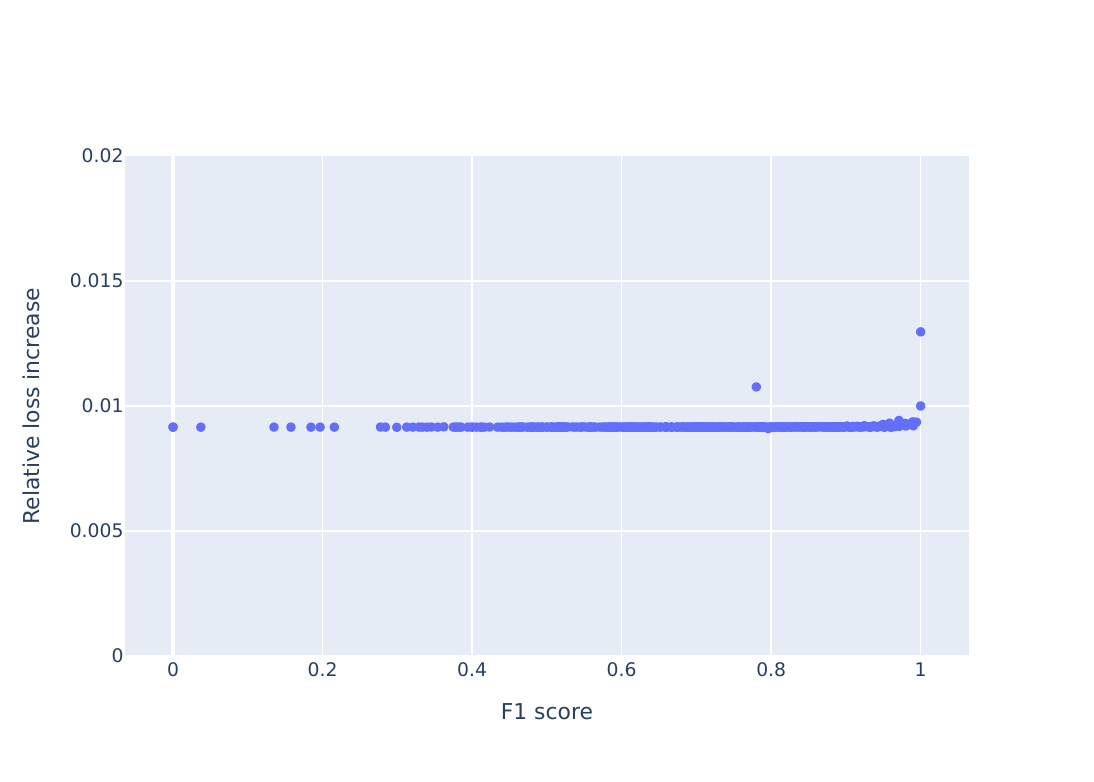}
    \caption{The effect on loss of ablating groups of features across the interpretability spectrum.  Every feature in a binary transcoder for layer 9 of a SmolLM2-135M is ordered by its fuzzing F1 score and assigned a bin such that each bin has an approximately equal firing count. Mean losses from zero ablating each bin are then taken over 1024-token sequences ($N=1024$.)}
    \label{fig:binned_ablation}
\end{figure*}

We also explored the relationship between the interpretability score and the causal effect of a feature on the model's performance. To do this, we grouped features into bins based on their fuzzing score, taking care to keep the total fire count roughly constant within each bin, and then ablated each bin of features and measured the resulting increase in cross-entropy loss. We found that almost all bins had nearly the same effect on the loss, except for two very high interpretability bins (Figure~\ref{fig:binned_ablation}). This is interesting and suggests that, at the extremes, there is a positive relationship between interpretability and causal influence. Across most bins, however, there is no relation between interpretability and causal importance.



\begin{table}[h]
\centering
\begin{tabular}{|c|ccccc|}
\hline
 & \multicolumn{5}{c|}{\textbf{Sparse probing ($\uparrow$)}} \\
\textbf{Layer} & SST & BSST & GSST & FSST & BFSST \\
\hline
L0 & 69.4 $\pm$ 1.7 & 64.5 $\pm$ 1.7 & \textbf{69.5 $\pm$ 1.0} & 69.1 $\pm$ 2.1 & 64.0 $\pm$ 0.5 \\
L9 & 60.3 $\pm$ 0.4 & 66.9 $\pm$ 3.5 & 67.6 $\pm$ 0.2 & 60.4 $\pm$ 0.7 & \textbf{67.8 $\pm$ 1.5} \\
L18 & 64.7 $\pm$ 2.1 & 68.2 $\pm$ 1.0 & 68.7 $\pm$ 1.2 & 66.4 $\pm$ 0.5 & \textbf{68.9 $\pm$ 0.9} \\
L27 & 74.7 $\pm$ 0.6 & 76.4 $\pm$ 0.8 & 76.8 $\pm$ 0.2 & 74.5 $\pm$ 0.5 & \textbf{77.3 $\pm$ 0.4} \\
\hline
\end{tabular}
\caption{Sparse probing evaluation from \cite{karvonen2024saebench}. Results for SmolLM2-135M for sparse skip-transcoders, binary sparse skip-transcoders, Gumbel binary sparse skip-transcoders, and finetuned variants of binary sparse skip-transcoders and sparse skip-transcoders. Binary sparse skip-transcoders are surprisingly useful in this setting, with both finetuned binary sparse coder variants scoring more highly than finetuned continuous sparse coders. Each sparse coder is evaluated over 3 seeds.}\label{tab:sparse_probing}
\end{table}

\subsection{Case study: a polysemantic code feature in SmolLM2-135M}

This feature (Figure~\ref{fig:topk_low_acts}) seems to fire inside doc strings in Python code in the top seven deciles of its activation range, although it can fire on a few different kinds of tokens within that context. The autointerp explanation is `"Special characters and symbols used for various purposes such as denoting comments, indicating code blocks, separating parameters, and representing mathematical operations."`.
In the lower deciles, it fires in a wider range of non-code contexts. We would like to eliminate the polysemanticity we see in the lower deciles. 


\subsection{Case study: the $search$ feature in SmolLM2-135M} \label{sec:search}
A non-cherrypicked feature was selected from the set of binary skip-transcoder features with perfect auto-interpretability fuzzing and detection scores (Fig.~\ref{fig:contrastive_examples}). The feature clearly activates on words related to searching (${search, searching, looking, quest, searched}$) but close inspection of the sequences containing activating examples reveals several non-activating instances of these words, both preceding and following the activating instances. This shows that our interpretability scores can over-estimate the true interpretability of features. This is partially because we use random non-activating examples when generating and evaluating explanations, rather than non-activating examples adversarially selected to be semantically similar to the activating ones. If the scoring pipeline had used adversarial selection of negative examples, it might have detected instances of the "search" concept where this feature does not fire.


\section{Discussion and future work}

While binary sparse coders produce more interpretable features, this effect is cancelled out by their tendency to produce uninterpretable ultra-high frequency features. They also explain less of the variance in model activations than continuous sparse coders, and they incur a larger increase in loss when patched into the model. Overall, we cannot recommend binary sparse coders for most purposes. The primary result of this work is negative: we attempted to remove a source of polysemanticity in sparse coding techniques (low activation values), and it re-appeared in a new form somewhere else (high-frequency features). This provides support for the idea that polysemanticity is an ineliminable feature of neural activations.

If polysemanticity turns out to be genuinely ineliminable, as we suspect but cannot prove, this would mean that the attempt to \emph{fully} decompose neural activations into interpretable features is fundamentally misguided. It would mean the mind is irreducibly holistic, rather than being built from logical atoms, as philosophers like Hubert Dreyfus have argued for decades \citep{dreyfus1976computers, dreyfus1992computers}.

More positively, future attempts to optimize the sparse coder architecture might include some mixture of binary and continuous features, since some concepts are best expressed as continuous values, and others discrete ones. While we could fix the proportion of binary features at the beginning of training as a hyperparameter, it may be best to dynamically determine the appropriate mixture in a data-dependent way. This might ameliorate the increase in fraction of variance unexplained we observe with binary sparse coders, and the problem of high-frequency features. Exploring mixed architectures is an interesting direction for future work. Notably, \citet{smith2025negative} investigated auxiliary loss terms that explicitly discourage the formation of high-frequency features, although they ultimately did not find the direction particularly promising.

Finally, the promising performance of binary sparse coders on sparse probing tasks could be investigated further.   

\section{Limitations}

Our focus on a single dataset may limit the applicability of this work---all sparse coders were trained on a sample of the SmolLM2 training corpus. This is a high-quality dataset, but using a different dataset could change the results. Additionally, some of our sparse coders exhibited high numbers of dead neurons (Figure~\ref{fig:16_index_collapse}). If this issue were resolved it might also change our results.

Our results may also change at higher sparse coder widths. Sparse coder FVU follows a scaling law, where increasing the number of features increases the fraction of variance explained. But as the width tends to infinity, the sparse coder converges to a degenerate solution where each individual input-output pair in the dataset gets its own feature. The optimal width for a sparse coder should make a tradeoff between reconstruction accuracy and learning features that are sufficiently coarse-grained to be interesting. The reduction of expressivity caused by binarising sparse coder activations could result in an optimal width that is too large to train with a reasonable amount of compute.



\bibliographystyle{plainnat}
\bibliography{citations}

\begin{thebibliography}{28}
\providecommand{\natexlab}[1]{#1}
\providecommand{\url}[1]{\texttt{#1}}
\expandafter\ifx\csname urlstyle\endcsname\relax
  \providecommand{\doi}[1]{doi: #1}\else
  \providecommand{\doi}{doi: \begingroup \urlstyle{rm}\Url}\fi

\bibitem[Ameisen et~al.(2025)Ameisen, Lindsey, Pearce, Gurnee, Turner, Chen, Citro, Abrahams, Carter, Hosmer, Marcus, Sklar, Templeton, Bricken, McDougall, Cunningham, Henighan, Jermyn, Jones, Persic, Qi, Ben~Thompson, Zimmerman, Rivoire, Conerly, Olah, and Batson]{ameisen2025circuit}
Emmanuel Ameisen, Jack Lindsey, Adam Pearce, Wes Gurnee, Nicholas~L. Turner, Brian Chen, Craig Citro, David Abrahams, Shan Carter, Basil Hosmer, Jonathan Marcus, Michael Sklar, Adly Templeton, Trenton Bricken, Callum McDougall, Hoagy Cunningham, Thomas Henighan, Adam Jermyn, Andy Jones, Andrew Persic, Zhenyi Qi, T.~Ben~Thompson, Sam Zimmerman, Kelley Rivoire, Thomas Conerly, Chris Olah, and Joshua Batson.
\newblock Circuit tracing: Revealing computational graphs in language models.
\newblock \emph{Transformer Circuits Thread}, 2025.
\newblock URL \url{https://transformer-circuits.pub/2025/attribution-graphs/methods.html}.

\bibitem[Arora et~al.(2018)Arora, Li, Liang, Ma, and Risteski]{arora2018linear}
Sanjeev Arora, Yuanzhi Li, Yingyu Liang, Tengyu Ma, and Andrej Risteski.
\newblock Linear algebraic structure of word senses, with applications to polysemy.
\newblock \emph{Transactions of the Association for Computational Linguistics}, 6:\penalty0 483--495, 2018.

\bibitem[Ayonrinde et~al.(2024)Ayonrinde, Pearce, and Sharkey]{ayonrinde2024interpretability}
Kola Ayonrinde, Michael~T Pearce, and Lee Sharkey.
\newblock Interpretability as compression: Reconsidering sae explanations of neural activations with mdl-saes.
\newblock \emph{arXiv preprint arXiv:2410.11179}, 2024.

\bibitem[Bengio et~al.(2013)Bengio, L{\'e}onard, and Courville]{bengio2013estimating}
Yoshua Bengio, Nicholas L{\'e}onard, and Aaron Courville.
\newblock Estimating or propagating gradients through stochastic neurons for conditional computation.
\newblock \emph{arXiv preprint arXiv:1308.3432}, 2013.

\bibitem[Bernstein et~al.(2018)Bernstein, Wang, Azizzadenesheli, and Anandkumar]{bernstein2018signsgd}
Jeremy Bernstein, Yu-Xiang Wang, Kamyar Azizzadenesheli, and Animashree Anandkumar.
\newblock signsgd: Compressed optimisation for non-convex problems.
\newblock In \emph{International Conference on Machine Learning}, pages 560--569. PMLR, 2018.

\bibitem[Bricken et~al.(2023)Bricken, Templeton, Batson, Chen, Jermyn, Conerly, Turner, Anil, Denison, Askell, Lasenby, Wu, Kravec, Schiefer, Maxwell, Joseph, Hatfield-Dodds, Tamkin, Nguyen, McLean, Burke, Hume, Carter, Henighan, and Olah]{bricken2023monosemanticity}
Trenton Bricken, Adly Templeton, Joshua Batson, Brian Chen, Adam Jermyn, Tom Conerly, Nick Turner, Cem Anil, Carson Denison, Amanda Askell, Robert Lasenby, Yifan Wu, Shauna Kravec, Nicholas Schiefer, Tim Maxwell, Nicholas Joseph, Zac Hatfield-Dodds, Alex Tamkin, Karina Nguyen, Brayden McLean, Josiah~E Burke, Tristan Hume, Shan Carter, Tom Henighan, and Christopher Olah.
\newblock Towards monosemanticity: Decomposing language models with dictionary learning.
\newblock \emph{Transformer Circuits Thread}, 2023.
\newblock https://transformer-circuits.pub/2023/monosemantic-features/index.html.

\bibitem[Bussmann et~al.(2024)Bussmann, Leask, and Nanda]{bussmann2024batchtopksparseautoencoders}
Bart Bussmann, Patrick Leask, and Neel Nanda.
\newblock Batchtopk sparse autoencoders.
\newblock \emph{arXiv preprint 2412.06410}, 2024.
\newblock URL \url{https://arxiv.org/abs/2412.06410}.

\bibitem[Defazio et~al.(2024)Defazio, Yang, Khaled, Mishchenko, Mehta, and Cutkosky]{defazio2024road}
Aaron Defazio, Xingyu Yang, Ahmed Khaled, Konstantin Mishchenko, Harsh Mehta, and Ashok Cutkosky.
\newblock The road less scheduled.
\newblock \emph{Advances in Neural Information Processing Systems}, 37:\penalty0 9974--10007, 2024.

\bibitem[Dreyfus(1976)]{dreyfus1976computers}
Hubert Dreyfus.
\newblock What computers can't do.
\newblock \emph{British Journal for the Philosophy of Science}, 27\penalty0 (2), 1976.

\bibitem[Dreyfus(1992)]{dreyfus1992computers}
Hubert~L Dreyfus.
\newblock \emph{What computers still can't do: A critique of artificial reason}.
\newblock MIT press, 1992.

\bibitem[Dunefsky et~al.(2024)Dunefsky, Chlenski, and Nanda]{dunefsky2024transcoders}
Jacob Dunefsky, Philippe Chlenski, and Neel Nanda.
\newblock Transcoders find interpretable llm feature circuits.
\newblock \emph{arXiv preprint arXiv:2406.11944}, 2024.

\bibitem[Elhage et~al.(2022)Elhage, Hume, Olsson, Schiefer, Henighan, Kravec, Hatfield-Dodds, Lasenby, Drain, Chen, et~al.]{elhage2022toy}
Nelson Elhage, Tristan Hume, Catherine Olsson, Nicholas Schiefer, Tom Henighan, Shauna Kravec, Zac Hatfield-Dodds, Robert Lasenby, Dawn Drain, Carol Chen, et~al.
\newblock Toy models of superposition.
\newblock \emph{arXiv preprint arXiv:2209.10652}, 2022.

\bibitem[Gallifant et~al.(2025)Gallifant, Chen, Sasse, Aerts, Hartvigsen, and Bitterman]{gallifant2025classification}
Jack Gallifant, Shan Chen, Kuleen Sasse, Hugo Aerts, Thomas Hartvigsen, and Danielle~S. Bitterman.
\newblock Sparse autoencoder features for classifications and transferability, 2025.
\newblock URL \url{https://arxiv.org/abs/2502.11367}.

\bibitem[Gao et~al.(2024)Gao, la~Tour, Tillman, Goh, Troll, Radford, Sutskever, Leike, and Wu]{gao2024scaling}
Leo Gao, Tom~Dupr{\'e} la~Tour, Henk Tillman, Gabriel Goh, Rajan Troll, Alec Radford, Ilya Sutskever, Jan Leike, and Jeffrey Wu.
\newblock Scaling and evaluating sparse autoencoders.
\newblock \emph{arXiv preprint arXiv:2406.04093}, 2024.

\bibitem[Guo et~al.(2025)Guo, Yang, Zhang, Song, Zhang, Xu, Zhu, Ma, Wang, Bi, et~al.]{guo2025deepseek}
Daya Guo, Dejian Yang, Haowei Zhang, Junxiao Song, Ruoyu Zhang, Runxin Xu, Qihao Zhu, Shirong Ma, Peiyi Wang, Xiao Bi, et~al.
\newblock Deepseek-r1: Incentivizing reasoning capability in llms via reinforcement learning.
\newblock \emph{arXiv preprint arXiv:2501.12948}, 2025.

\bibitem[Gurnee et~al.(2023)Gurnee, Nanda, Pauly, Harvey, Troitskii, and Bertsimas]{gurnee2023finding}
Wes Gurnee, Neel Nanda, Matthew Pauly, Katherine Harvey, Dmitrii Troitskii, and Dimitris Bertsimas.
\newblock Finding neurons in a haystack: Case studies with sparse probing.
\newblock \emph{arXiv preprint arXiv:2305.01610}, 2023.

\bibitem[Gurnee et~al.(2024)Gurnee, Horsley, Guo, Kheirkhah, Sun, Hathaway, Nanda, and Bertsimas]{gurnee2024universal}
Wes Gurnee, Theo Horsley, Zifan~Carl Guo, Tara~Rezaei Kheirkhah, Qinyi Sun, Will Hathaway, Neel Nanda, and Dimitris Bertsimas.
\newblock Universal neurons in gpt2 language models.
\newblock \emph{arXiv preprint arXiv:2401.12181}, 2024.

\bibitem[Jang et~al.(2017)Jang, Gu, and Poole]{jang2017categorical}
Eric Jang, Shixiang Gu, and Ben Poole.
\newblock Categorical reparameterization with gumbel-softmax.
\newblock In \emph{International Conference on Learning Representations}, 2017.
\newblock URL \url{https://openreview.net/forum?id=rkE3y85ee}.

\bibitem[Karvonen(2025)]{karvonen2025finetuneSAE}
Adam Karvonen.
\newblock Revisiting end-to-end sparse autoencoder training: A short finetune is all you need, 2025.
\newblock URL \url{https://arxiv.org/abs/2503.17272}.

\bibitem[Karvonen et~al.(2024)Karvonen, Rager, Lin, Tigges, Bloom, Chanin, Lau, Farrell, Conmy, McDougall, Ayonrinde, Wearden, Marks, and Nanda]{karvonen2024saebench}
Adam Karvonen, Can Rager, Johnny Lin, Curt Tigges, Joseph Bloom, David Chanin, Yeu-Tong Lau, Eoin Farrell, Arthur Conmy, Callum McDougall, Kola Ayonrinde, Matthew Wearden, Samuel Marks, and Neel Nanda.
\newblock Saebench: A comprehensive benchmark for sparse autoencoders, 2024.
\newblock URL \url{https://www.neuronpedia.org/sae-bench/info}.
\newblock Accessed: 2025-01-17.

\bibitem[Kingma and Ba(2017)]{kingma2017adam}
Diederik~P. Kingma and Jimmy Ba.
\newblock Adam: A method for stochastic optimization, 2017.
\newblock URL \url{https://arxiv.org/abs/1412.6980}.

\bibitem[Olah et~al.(2020)Olah, Cammarata, Schubert, Goh, Petrov, and Carter]{Olah2020}
Chris Olah, Nick Cammarata, Ludwig Schubert, Gabriel Goh, Michael Petrov, and Shan Carter.
\newblock Zoom in: An introduction to circuits.
\newblock \emph{Distill}, 5\penalty0 (3), March 2020.
\newblock ISSN 2476-0757.
\newblock \doi{10.23915/distill.00024.001}.
\newblock URL \url{http://dx.doi.org/10.23915/distill.00024.001}.

\bibitem[Paulo et~al.(2024)Paulo, Mallen, Juang, and Belrose]{paulo2024delphi}
Gonçalo Paulo, Alex Mallen, Caden Juang, and Nora Belrose.
\newblock Automatically interpreting millions of features in large language models, 2024.
\newblock URL \url{https://arxiv.org/abs/2410.13928}.

\bibitem[Paulo et~al.(2025)Paulo, Shabalin, and Belrose]{paulo2025transcoders}
Gonçalo Paulo, Stepan Shabalin, and Nora Belrose.
\newblock Transcoders beat sparse autoencoders for interpretability, 2025.
\newblock URL \url{https://arxiv.org/abs/2501.18823}.

\bibitem[Rajamanoharan et~al.(2024)Rajamanoharan, Lieberum, Sonnerat, Conmy, Varma, Kram{\'a}r, and Nanda]{rajamanoharan2024jumping}
Senthooran Rajamanoharan, Tom Lieberum, Nicolas Sonnerat, Arthur Conmy, Vikrant Varma, J{\'a}nos Kram{\'a}r, and Neel Nanda.
\newblock Jumping ahead: Improving reconstruction fidelity with jumprelu sparse autoencoders.
\newblock \emph{arXiv preprint arXiv:2407.14435}, 2024.

\bibitem[Smith et~al.(2025)Smith, Rajamanoharan, Conmy, McDougall, Kramar, Lieberum, Shah, and Nanda]{smith2025negative}
Lewis Smith, Sen Rajamanoharan, Arthur Conmy, Callum McDougall, Janos Kramar, Tom Lieberum, Rohin Shah, and Neel Nanda.
\newblock Negative results for sparse autoencoders on downstream tasks and deprioritising sae research (mechanistic interpretability team progress update), March 2025.
\newblock URL \url{https://deepmindsafetyresearch.medium.com/negative-results-for-sparse-autoencoders-on-downstream-tasks-and-deprioritising-sae-research-6cadcfc125b9}.

\bibitem[Tamkin et~al.(2023)Tamkin, Taufeeque, and Goodman]{tamkin2023codebook}
Alex Tamkin, Mohammad Taufeeque, and Noah~D. Goodman.
\newblock Codebook features: Sparse and discrete interpretability for neural networks, 2023.
\newblock URL \url{https://arxiv.org/abs/2310.17230}.

\bibitem[Templeton et~al.(2024)Templeton, Conerly, Marcus, Lindsey, Bricken, Chen, Pearce, Citro, Ameisen, Jones, Cunningham, Turner, McDougall, MacDiarmid, Freeman, Sumers, Rees, Batson, Jermyn, Carter, Olah, and Henighan]{templeton2024scaling}
Adly Templeton, Tom Conerly, Jonathan Marcus, Jack Lindsey, Trenton Bricken, Brian Chen, Adam Pearce, Craig Citro, Emmanuel Ameisen, Andy Jones, Hoagy Cunningham, Nicholas~L Turner, Callum McDougall, Monte MacDiarmid, C.~Daniel Freeman, Theodore~R. Sumers, Edward Rees, Joshua Batson, Adam Jermyn, Shan Carter, Chris Olah, and Tom Henighan.
\newblock Scaling monosemanticity: Extracting interpretable features from claude 3 sonnet.
\newblock \emph{Transformer Circuits Thread}, 2024.
\newblock URL \url{https://transformer-circuits.pub/2024/scaling-monosemanticity/index.html}.

\end{thebibliography}


\appendix

\newpage
\section{Technical Appendices and Supplementary Material}


\subsection{Compute}


Each experiment was run on nodes equipped with either 4 or 8 GPUs. The nodes were equipped with a mixture of NVIDIA A40, A100, and H100 GPUs from various cloud providers. Individual training runs took 3 to 15 hours, full-layer binned ablations took 4 hours, and collecting individual model interpretability statistics took around 1 hour per model. The total compute estimate is 500 A100 compute hours. Additionally, preliminary experiments used an estimated 100 A100 compute hours.

\subsection{Results for SmolLM2-1.7B binary and continuous skip-transcoders}

\begin{table}[h]
\centering
\begin{tabular}{|c|cc|}
\hline
 & \multicolumn{2}{c|}{\textbf{SmolLM2-1.7B TC}} \\
\textbf{Metric} & Binary & Continuous \\ \hline
Unweighted fuzzing F1     & \textbf{0.880} & 0.779 \\
Unweighted detection F1   & \textbf{0.8387}  & 0.670 \\ 
Weighted fuzzing F1     & \textbf{0.801} & 0.658 \\
Weighted detection F1   & \textbf{0.702}  & 0.670 \\ \hline
\end{tabular}
\caption{Results for SmolLM2-1.7B binary and continuous skip-transcoders, expansion factor of 64, $k=32$, averaged over four layers. In this single-seed run BTCs are consistently more interpretable.}
\end{table}

\subsection{Detection F1 scores}

\begin{figure*}[h]
    \centering
    \includegraphics[trim=0 0 0 0, clip, width=1.\textwidth]{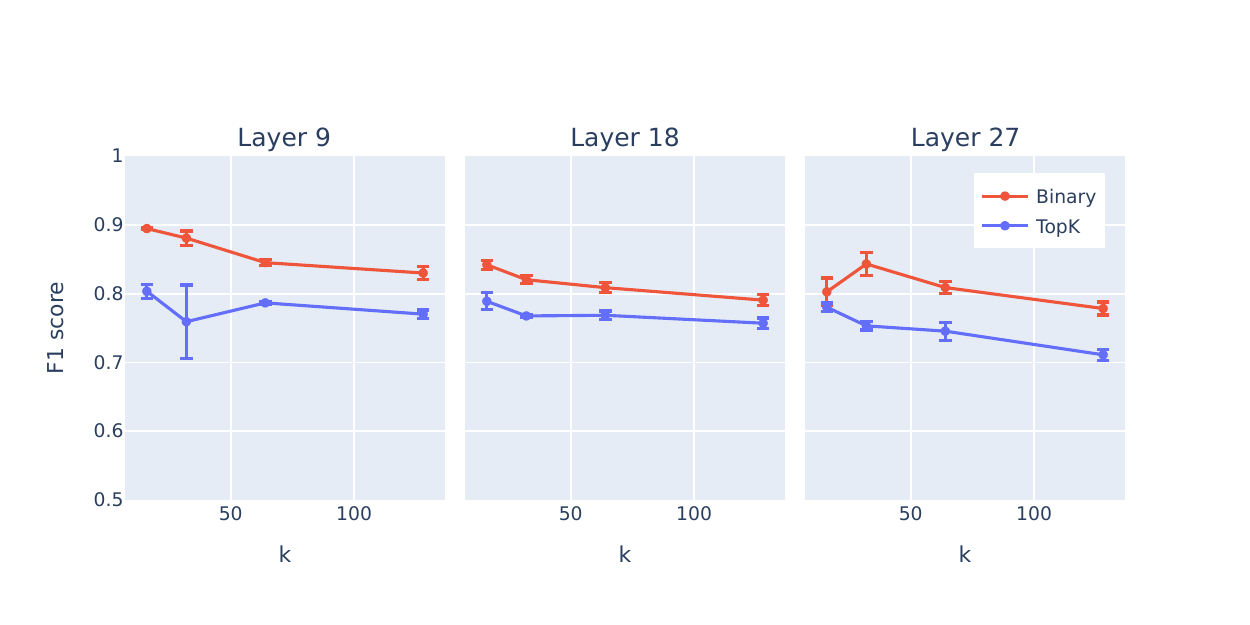}
    \caption{Unweighted detection F1 scores for SmolLM2-135M binary and continuous skip-transcoders.}
    \label{fig:detection_btc}
\end{figure*}

\begin{figure*}[h]
    \centering
    \includegraphics[trim=0 0 0 0, clip, width=1.\textwidth]{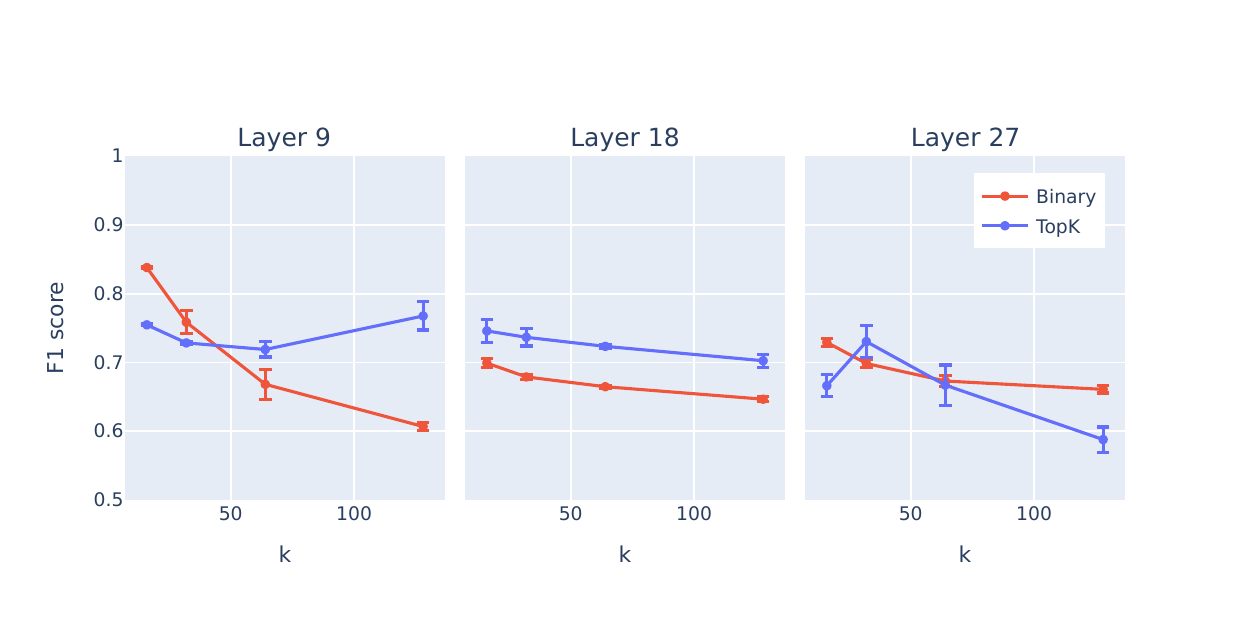}
    \caption{Unweighted detection F1 scores for SmolLM2-135M BAEs and SAEs.}
    \label{fig:detection_bae_unweighted}
\end{figure*}

\begin{figure*}[h]
    \centering
    \includegraphics[trim=0 0 0 0, clip, width=1.\textwidth]{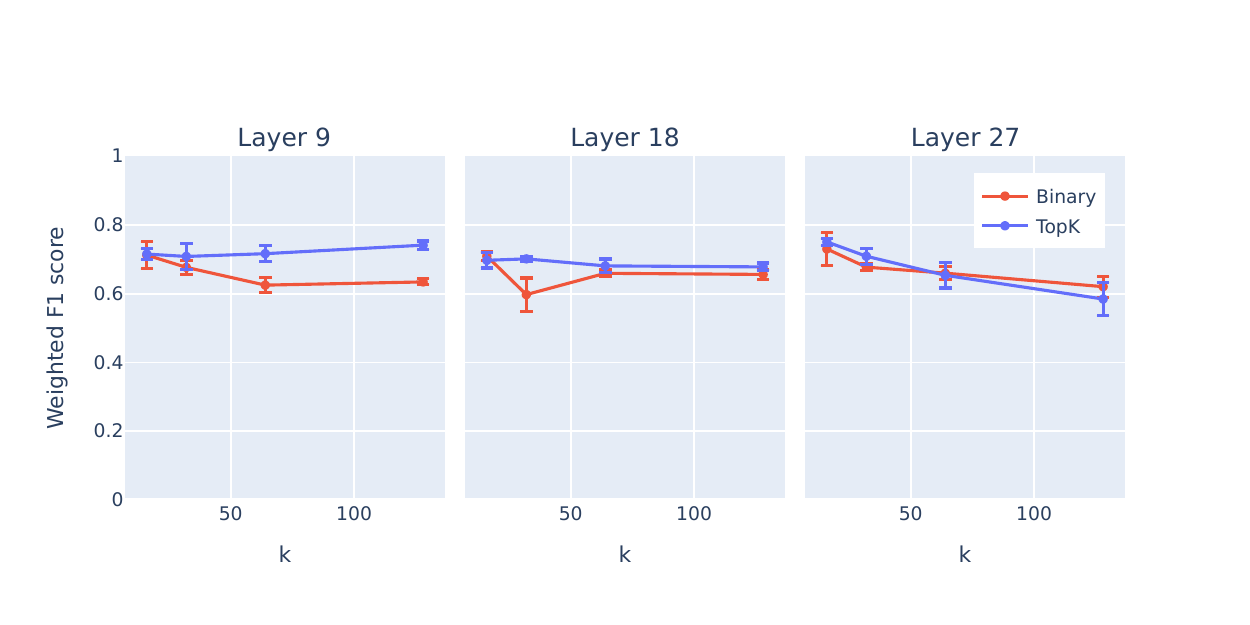}
    \caption{Frequency-weighted detection F1 scores for SmolLM2-135M binary and continuous skip-transcoders.}
    \label{fig:detection_btc_weighted}
\end{figure*}

\begin{figure*}[h]
    \centering
    \includegraphics[trim=0 0 0 0, clip, width=1.\textwidth]{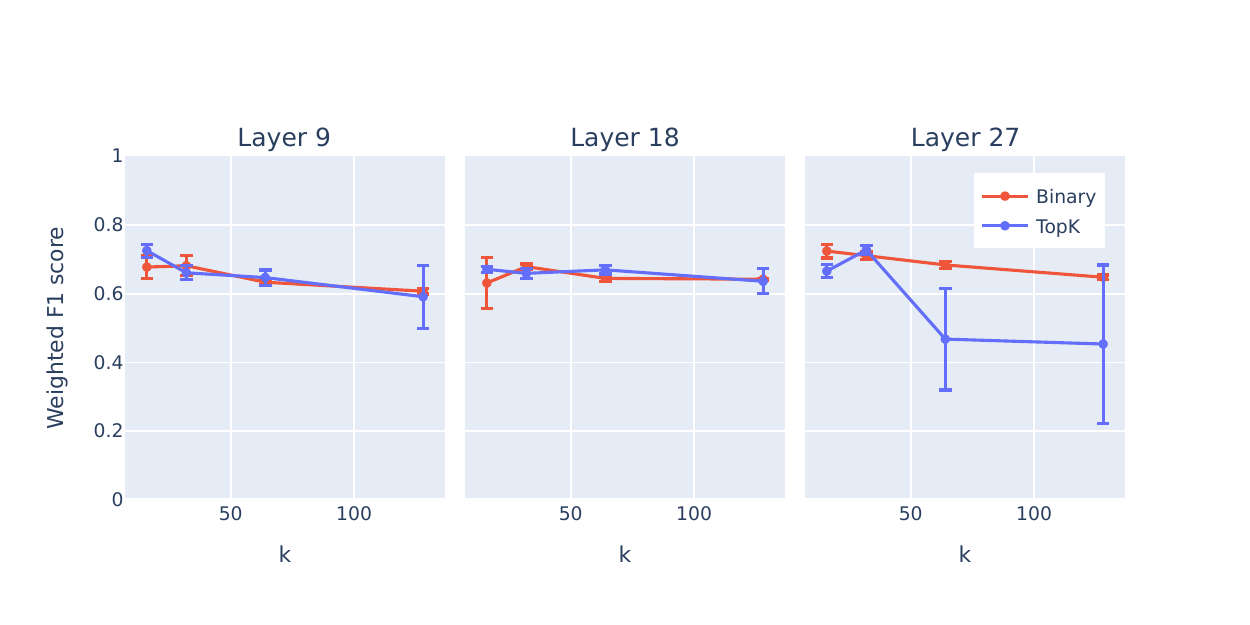}
    \caption{Frequency-weighted detection F1 scores for SmolLM2-135M BAEs and SAEs.}
    \label{fig:detection_bae_weighted}
\end{figure*}

\clearpage

\subsection{Fraction of variance unexplained}\label{app:fvu}

\begin{figure*}[h]
    \centering
    \includegraphics[trim=0 0 0 0, clip, width=0.6\textwidth]{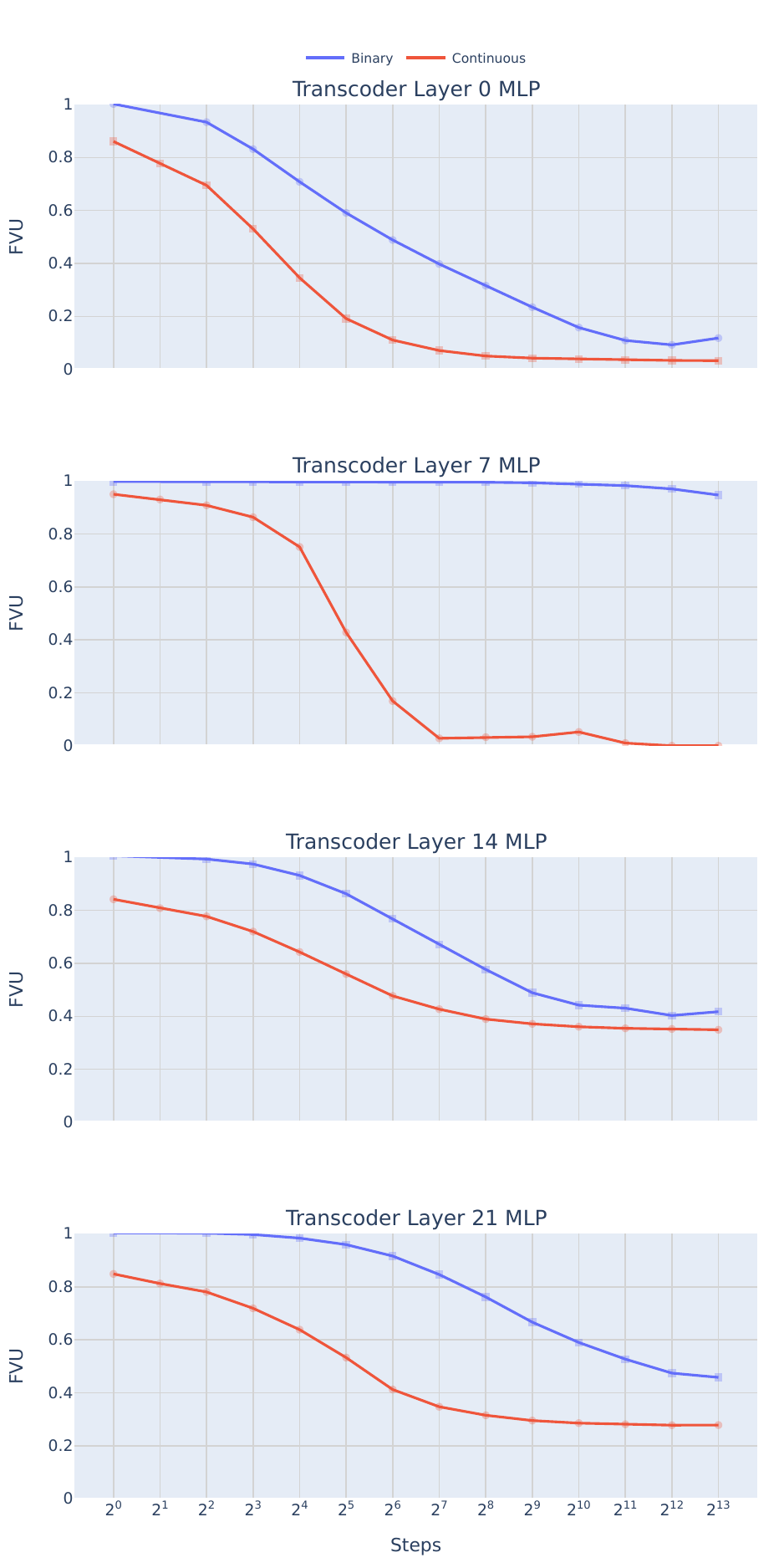}
    \caption{FVUs over training for SmolLM2-1.7B binary and continuous skip-transcoders.}
    \label{fig:fvu_1_7b}
\end{figure*}

\begin{figure*}[h]
    \centering
    \includegraphics[trim=0 0 0 0, clip, width=1.\textwidth]{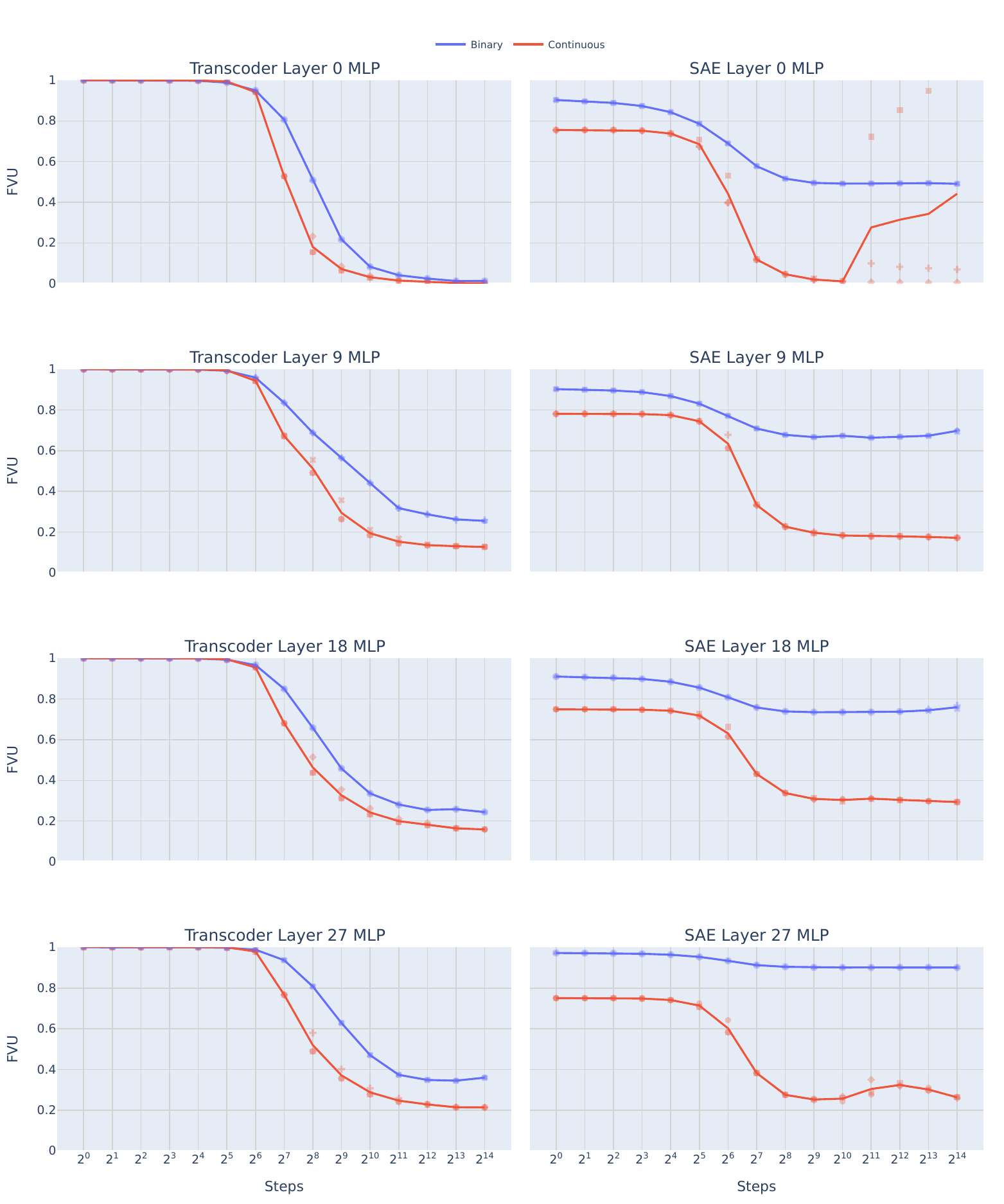}
    \caption{FVUs over training for all $k=16$ pre-trained models.}
    \label{fig:fvu_16}
\end{figure*}

\begin{figure*}[h]
    \centering
    \includegraphics[trim=0 0 0 0, clip, width=1.\textwidth]{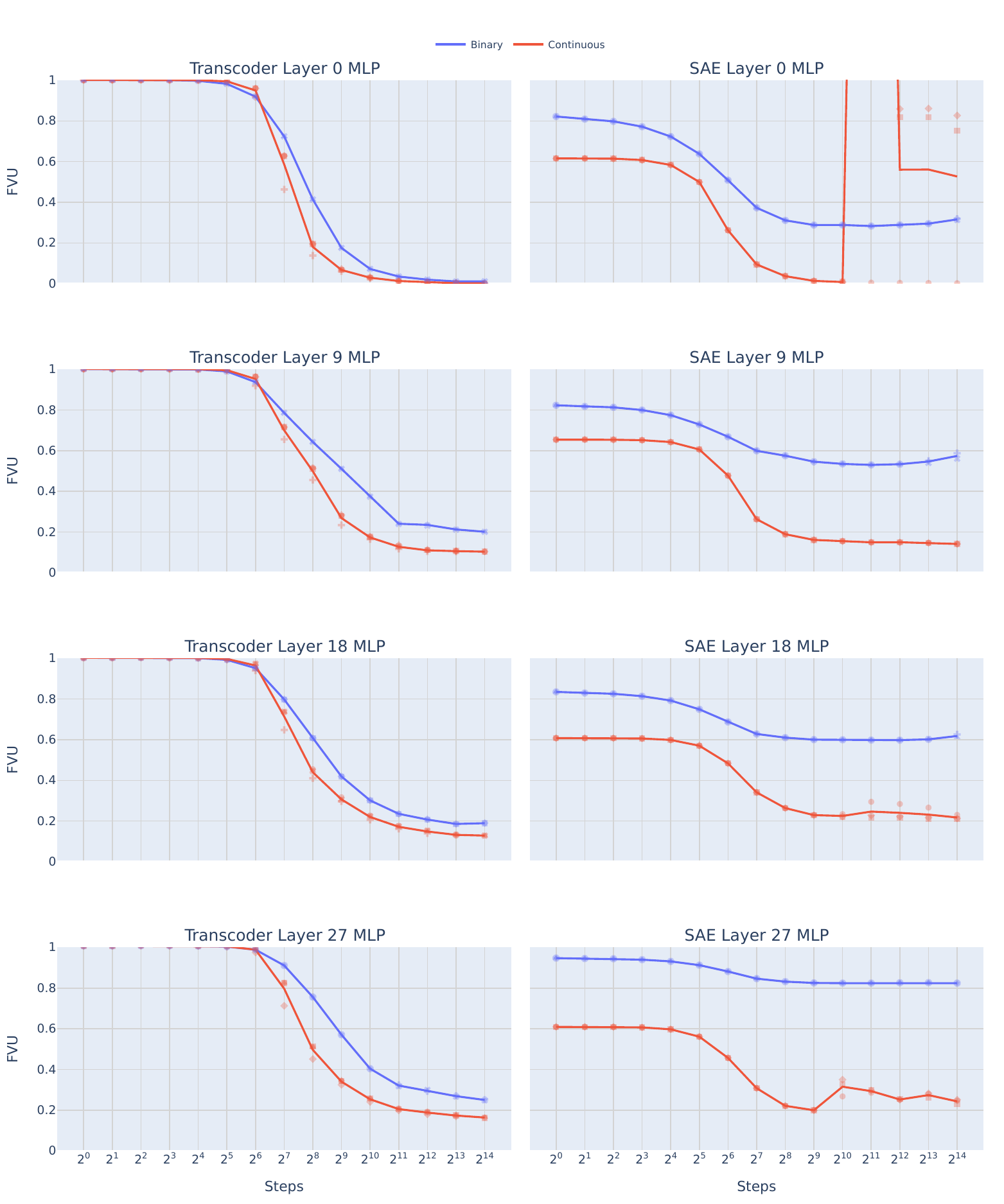}
    \caption{FVUs over training for all $k=32$ pre-trained models.}
    \label{fig:fvu_32}
\end{figure*}

\begin{figure*}[h]
    \centering
    \includegraphics[trim=0 0 0 0, clip, width=1.\textwidth]{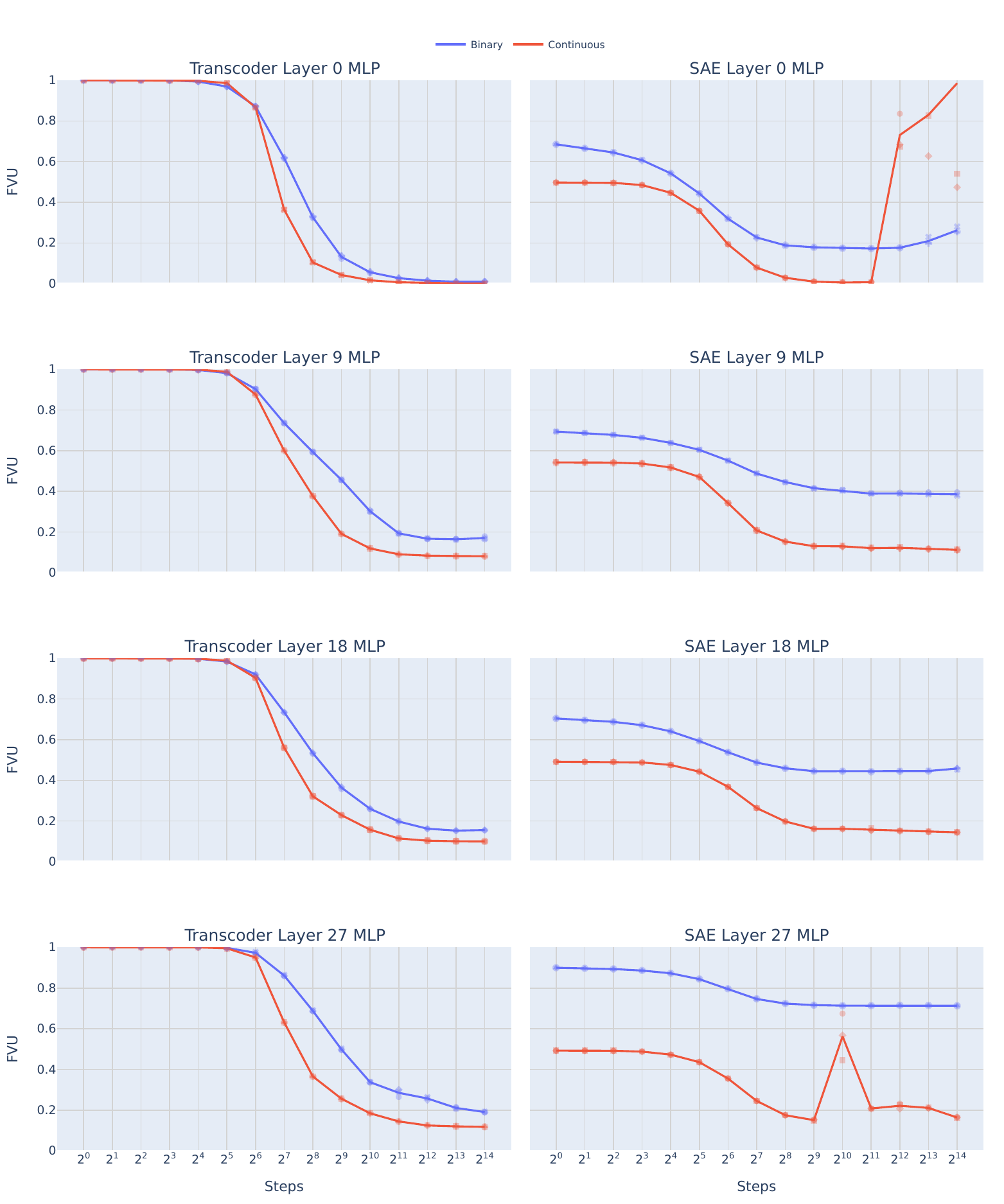}
    \caption{FVUs over training for all $k=64$ pre-trained models.}
    \label{fig:fvu_64}
\end{figure*}

\begin{figure*}[h]
    \centering
    \includegraphics[trim=0 0 0 0, clip, width=1.\textwidth]{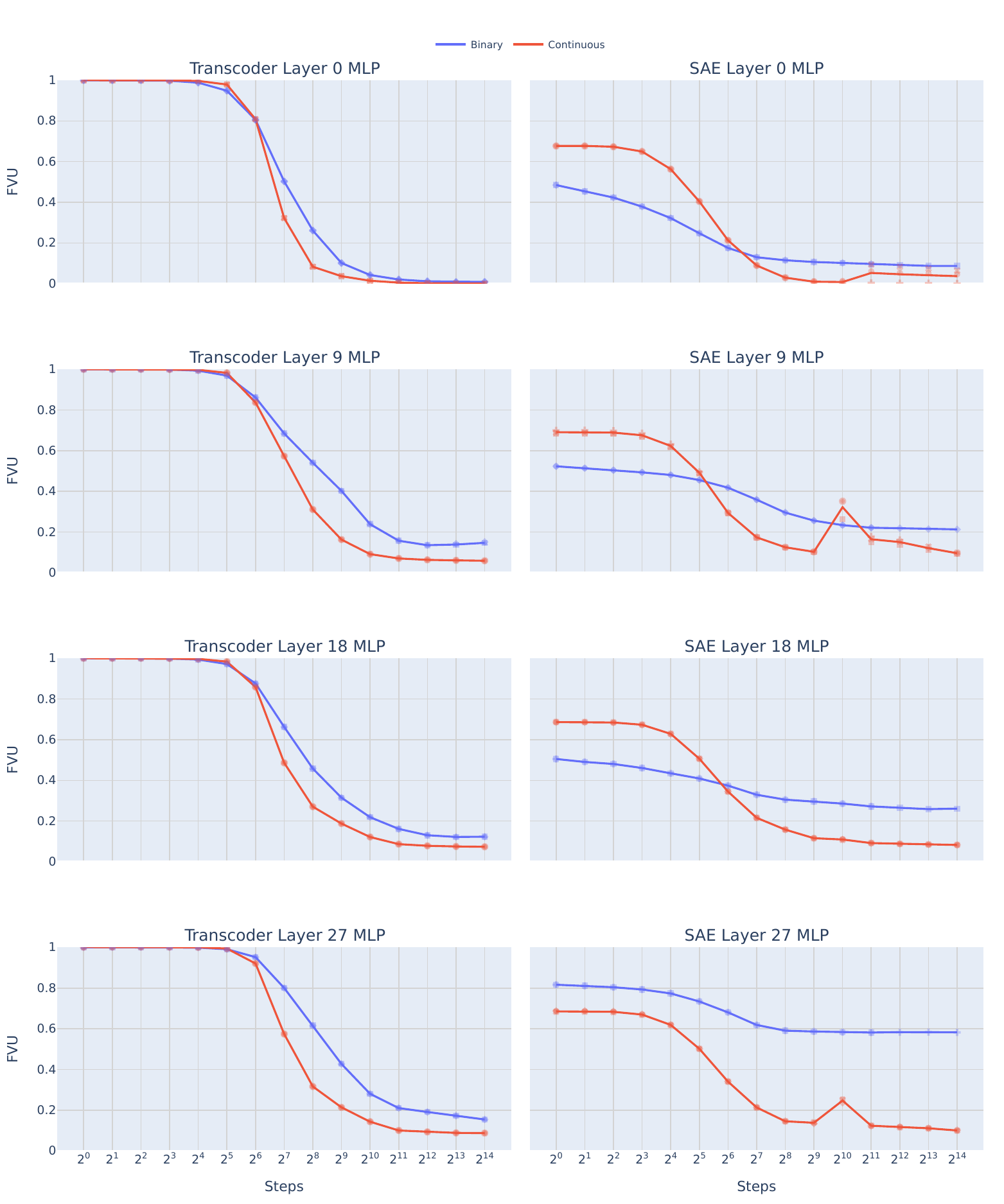}
    \caption{FVUs over training for all $k=128$ pre-trained models.}
    \label{fig:fvu_128}
\end{figure*}

\begin{figure*}[h]
    \centering
    \includegraphics[trim=0 0 0 0, clip, width=0.8\textwidth]{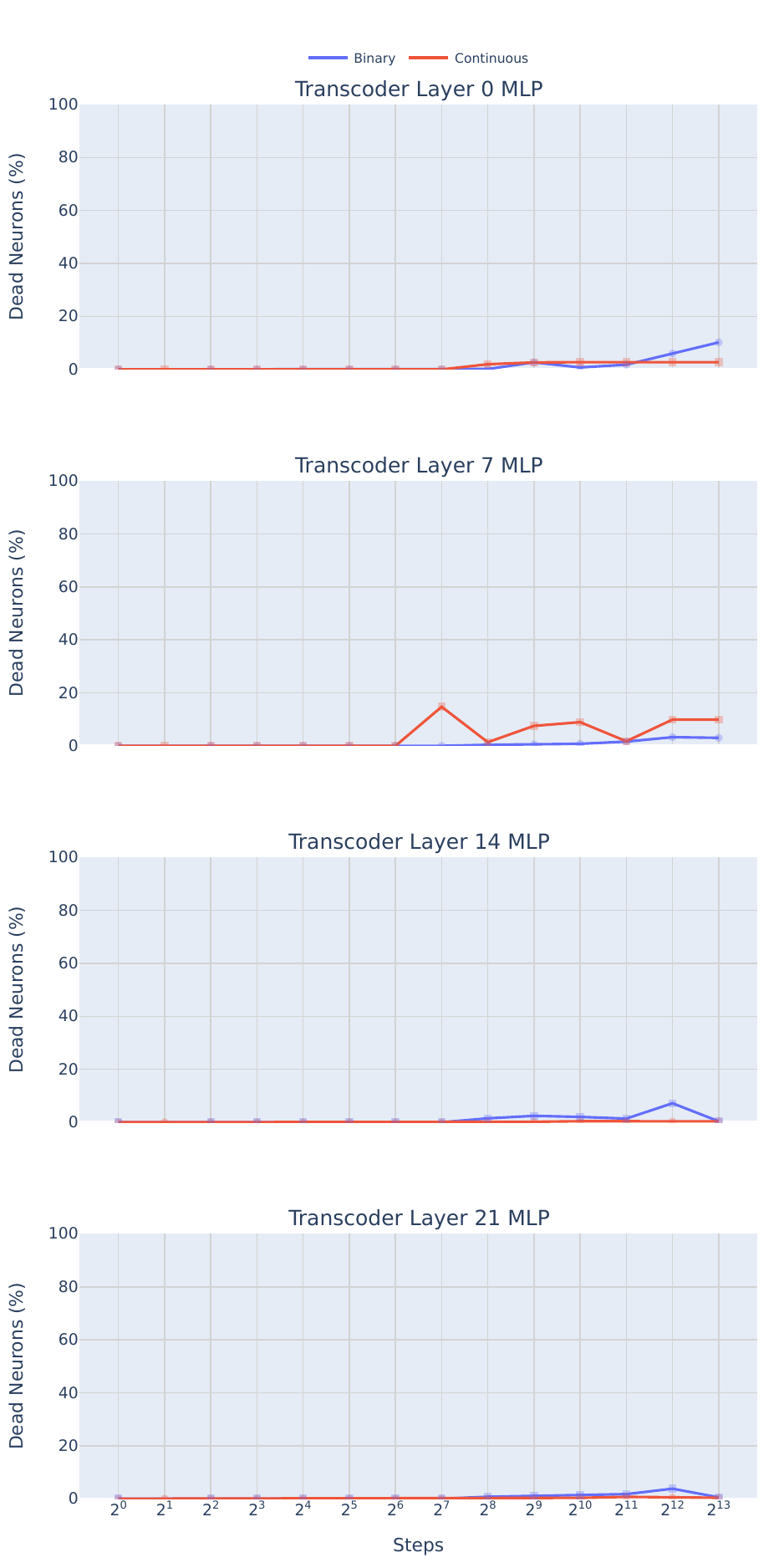}
    \caption{Dead neurons over training for SmolLM2-1.7B binary and continuous skip-transcoders.}
    \label{fig:1_7b_index_collapse}
\end{figure*}

\begin{figure*}[h]
    \centering
    \includegraphics[trim=0 0 0 0, clip, width=1.\textwidth]{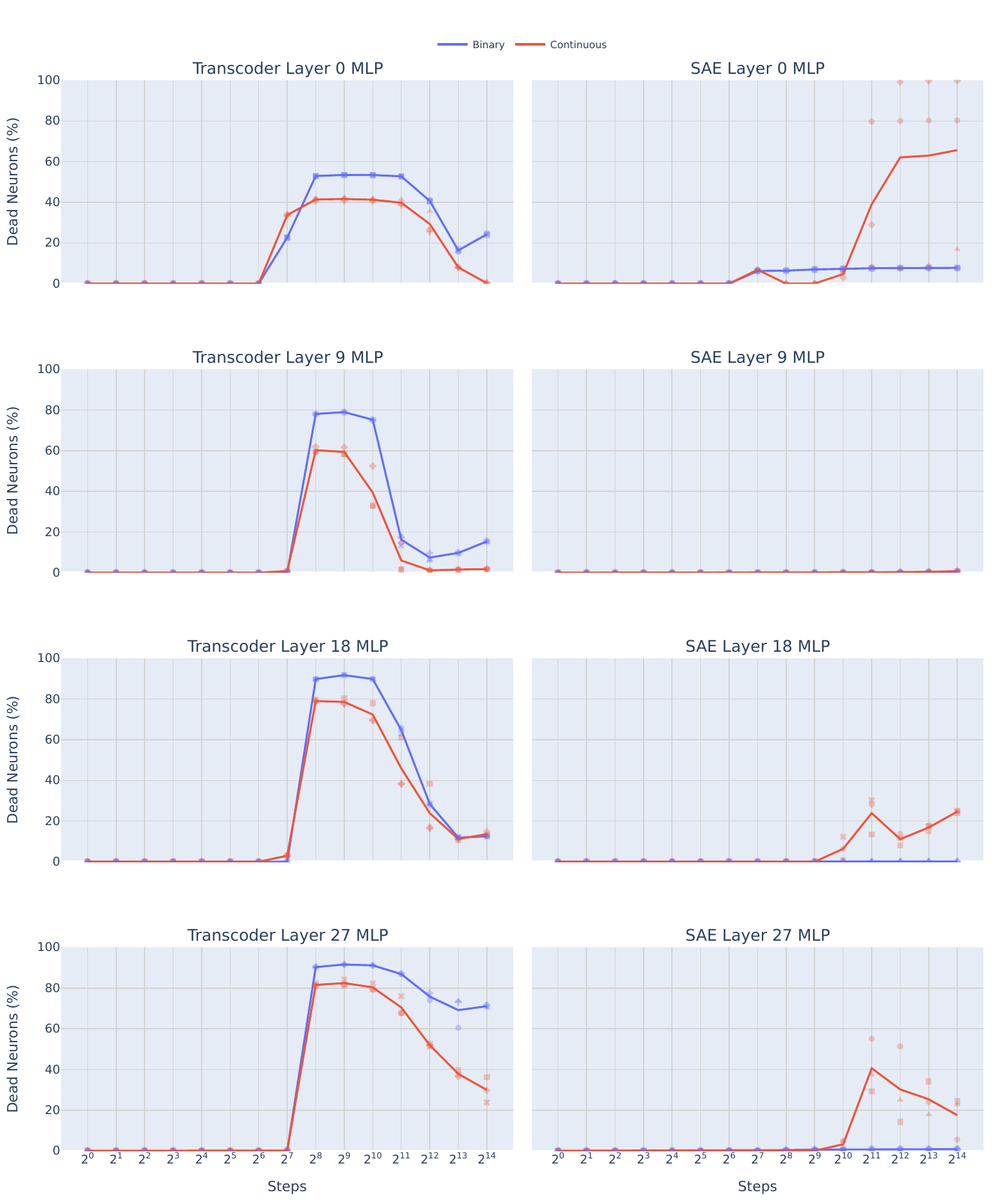}
    \caption{Dead neurons over training for all $k=16$ pre-trained models.}
    \label{fig:16_index_collapse}
\end{figure*}

\begin{figure*}[h]
    \centering
    \includegraphics[trim=0 0 0 0, clip, width=1.\textwidth]{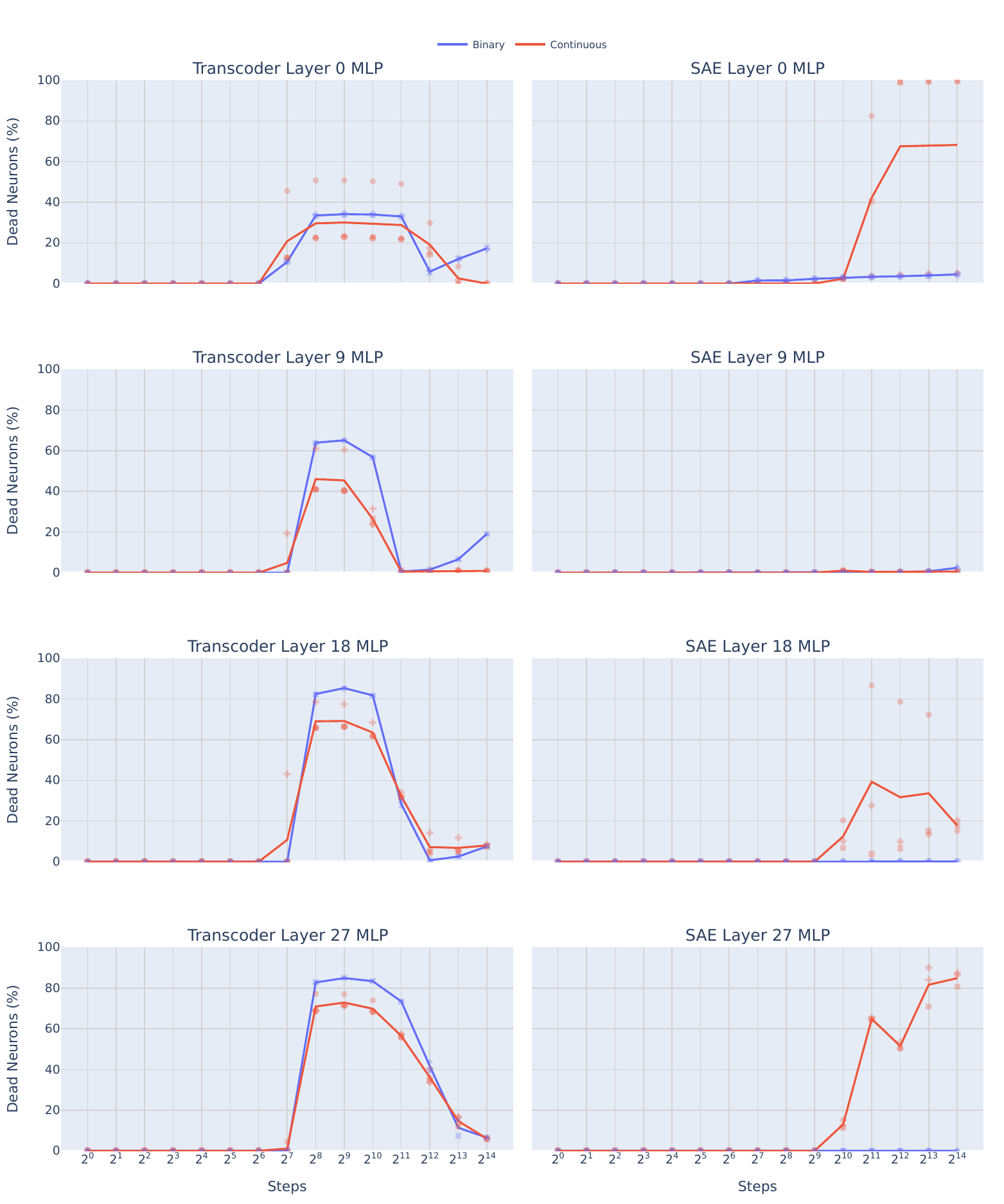}
    \caption{Dead neurons over training for all $k=32$ pre-trained models.}
    \label{fig:32_index_collapse}
\end{figure*}

\begin{figure*}[h]
    \centering
    \includegraphics[trim=0 0 0 0, clip, width=1.\textwidth]{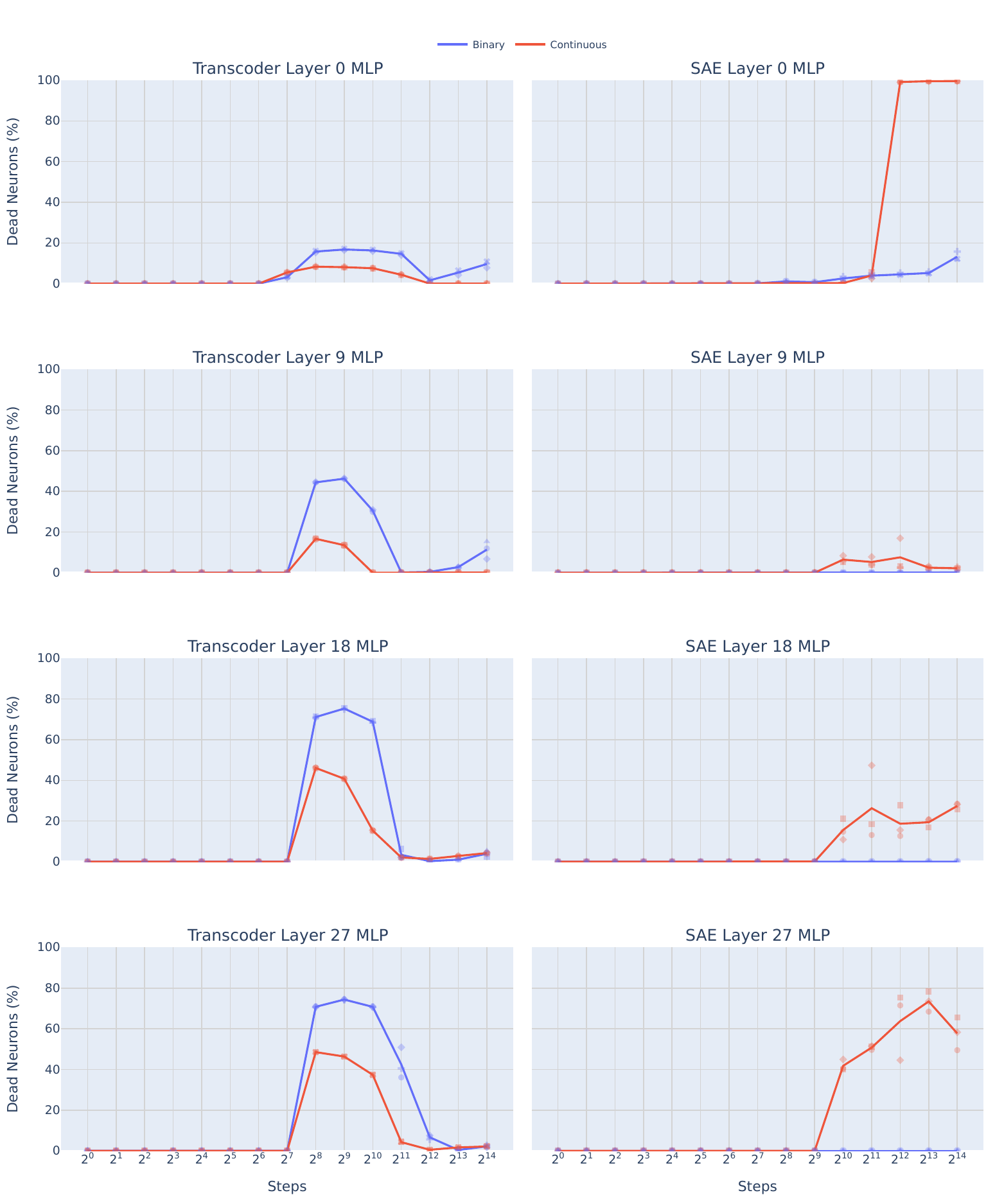}
    \caption{Dead neurons over training for all $k=64$ pre-trained models.}
    \label{fig:64_index_collapse}
\end{figure*}

\begin{figure*}[h]
    \centering
    \includegraphics[trim=0 0 0 0, clip, width=1.\textwidth]{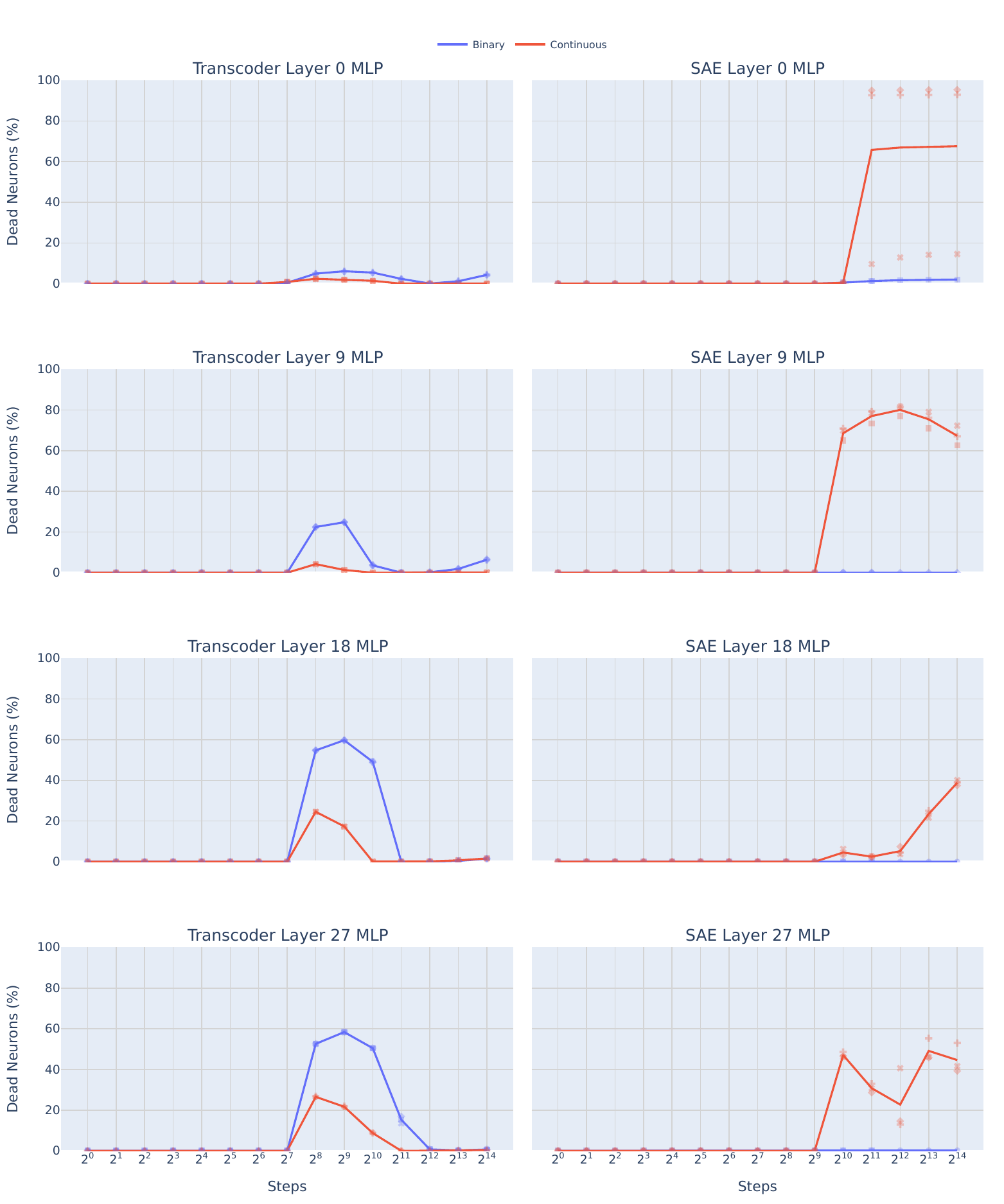}
    \caption{Dead neurons over training for all $k=128$ pre-trained models.}
    \label{fig:128_index_collapse}
\end{figure*}


\clearpage

\end{document}